\documentclass[11pt]{article}

\pdfoutput=1
\usepackage[final]{acl}

\usepackage{times}
\usepackage{latexsym}

\usepackage[T1]{fontenc}
\usepackage[utf8]{inputenc}

\usepackage{microtype}

\usepackage{inconsolata}

\usepackage{graphicx}
\usepackage{authblk}
\usepackage{booktabs} 
\usepackage{subfiles}
\usepackage{amsmath}
\usepackage{subcaption}
\usepackage{makecell}
\usepackage{tabularx}
\usepackage{array}
\usepackage{adjustbox}
\usepackage{geometry}
\usepackage{multicol}
\usepackage{multirow}
\usepackage{longtable}
\usepackage{hyperref}
\usepackage{listings}
\usepackage{adjustbox}

\title{\textit{Fill In The Gaps}: Model Calibration and Generalization with Synthetic Data}

\author[1]{Yang Ba}
\author[2]{Michelle V. Mancenido}
\author[1]{Rong Pan}

\affil[1]{School of Computing and Augmented Intelligence, Arizona State University}
\affil[2]{School of Mathematical and Natural Sciences, Arizona State University}

\affil[ ]{\normalsize\textsuperscript{1}\href{mailto:yangba@asu.edu}{yangba@asu.edu}, \href{mailto:Rong.Pan@asu.edu}{Rong.Pan@asu.edu}, \textsuperscript{2}\href{mailto:mmanceni@asu.edu}{mmanceni@asu.edu}}

\begin{document}

\maketitle

\begin{abstract}
As machine learning models continue to swiftly advance, calibrating their performance has become a major concern prior to practical and widespread implementation. Most existing calibration methods often negatively impact model accuracy due to the lack of diversity of validation data, resulting in reduced generalizability. To address this, we propose a calibration method that incorporates synthetic data without compromising accuracy. We derive the expected calibration error (ECE) bound using the Probably Approximately Correct (PAC) learning framework. Large language models (LLMs), known for their ability to mimic real data and generate text with mixed class labels, are utilized as a synthetic data generation strategy to lower the ECE bound and improve model accuracy on real test data. Additionally, we propose data generation mechanisms for efficient calibration. Testing our method on four different natural language processing tasks, we observed an average up to 34\% increase in accuracy and 33\% decrease in ECE.

 % 1. calibration and generalization 
 % 2. synthetic data with LLMs with minimal cost 
 % 3. use syhtentic data to fill the gaps to achieve both 
 % 4. we apply out approach in XXX datasets and observe xx\% increase , xx\% decrease . We also compare our methods on ... 
 %  goal: Improve calibration without sacrificing performance (more data should improve performance on test set) - edit later 
 %  xx\% increase , xx\% decrease  
 %  test Because of the popularity of generative models, high-quality synthetic data can be obtained with a minimal cost, which makes synthetic data an good option to drive the algorithms to be responsible and robust. 

\end{abstract}

% \section{Introduction}
% \subfile{sections/1.introduction}

% \section{Calibration Concept}
% \subfile{sections/2.preliminary}

% \section{Methodology}
% \subfile{sections/3.methodology}

% \section{Experiment}
% \subfile{sections/4.experiment}

% \section{Related Work} 
% \subfile{sections/5.related work}

% \section{Conclusion} % or discussion
% \subfile{sections/6.conclusion}

% \section*{Limitations and Future Work}
% \subfile{sections/7.limitation}

\section{Introduction}

% \section{Introduction}

% "On Calibration of Modern Neural Networks" states high accuracy and well-calibrated model cannot be achieve at the same time. (see figures in the paper)  

% P1: Text classaction task, high performance, - > calibration, different tasks specifically 
Natural Language Processing (NLP) models have fundamentally advanced the syntactic and semantic analysis, information retrieval, and automated generation of textual data. State-of-the-art (SOTA) models (e.g., transformers \citep{vaswani2017attention}, BERT \cite{devlin2018bert}, and RoBERTa \cite{liu2019roberta}) have excelled in practical, user-centric applications such as automated customer support chatbots, personalized content curation, and real-time multilingual text translation. Other NLP models, which are typically trained for a specialized use context, have also been developed and fine-tuned for numerous downstream tasks, including sentiment analysis, named entity recognition (NER), and text classification, as parts of a decision-support system (DSS). Powered by deep learning algorithms, these classification models have achieved remarkable levels of performance in terms of their accuracy, F1 scores, and AUCs \cite{li2020deep, cohan2019pretrained}. 

As machine learning philosophies continue to evolve, growing attention is placed on metrics beyond simple classification accuracy. In recent years, socially responsible artificial intelligence (AI) has been strongly advocated by algorithmic regulatory frameworks (e.g., the US Algorithmic Accountability Act \cite{donovan2018algorithmic}), especially in safety-critical domains, such as healthcare \cite{pfohl2022net} and law enforcement \cite{salvador2021faircal}. Some key pillars of socially responsible AI include \emph{accountability}, \emph{transparency}, and \emph{robustness} \cite{cooper2022accountability}. Ensuring a calibrated ML model accountable for its decision means that it must provide clear justifications for any decision being made, while transparency requires that these justifications are understandable and interpretable \cite{kadavath2022language}; additionally, robustness requires that the ML model performs consistently well under various conditions. In classification tasks, these requirements can be addressed by properly managing model output uncertainty, i.e., quantifying, calibrating, and communicating the proper confidence level associated with each prediction to the end user. Among the three aspects of uncertainty management, calibration directly improves model performance by ensuring that model predictions are congruent with empirically observed outcomes. 

% \begin{figure*}[t]
% \centering
% \begin{adjustbox}
%  \includegraphics[width=1.0\textwidth]{figures/calibration.final.png}
% \end{adjustbox}
% \caption{\label{fig:framework} The framework of our proposed method involves initially training a downstream model using real data to identify poorly calibrated bins. Data instances, $x_{i}$, and their prediction probabilities, $\hat{P}(y_{i}|x_{i})$, from these bins are then fed into large language models (LLMs) to generate synthetic data, $x^{syn}_{i}$, along with their corresponding probabilities, $\hat{P}(y_{i}|x^{syn}_{i})$. This synthetic data, combined with the real data, is used to retrain the downstream model, thereby improving calibration outputs without compromising model performance.}  
% \end{figure*}

\begin{figure*}[t]
    \centering
    \begin{adjustbox}{width=1.0\textwidth} % Set the width of the adjustbox
        \includegraphics{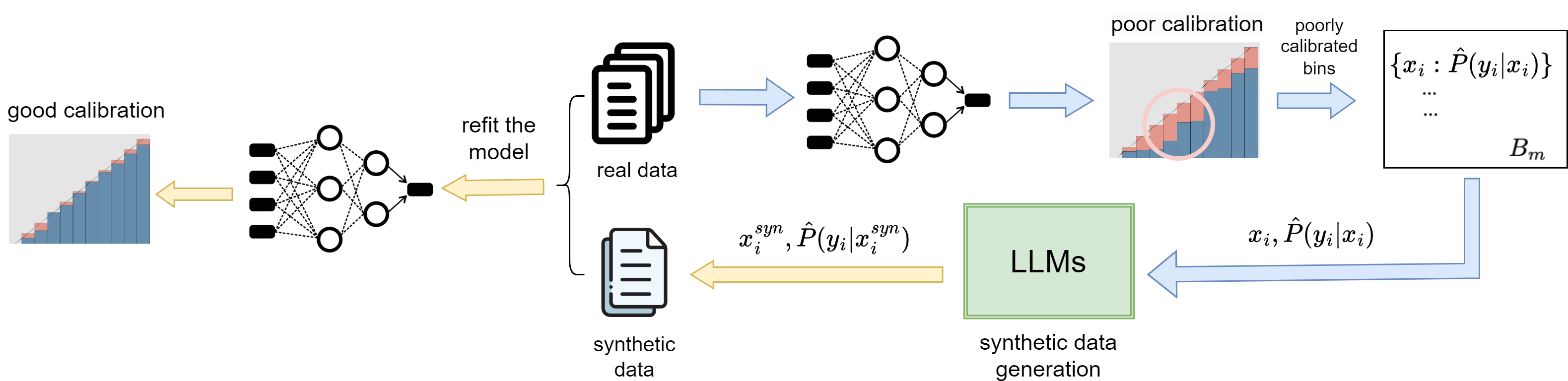} % Adjust the path as necessary
    \end{adjustbox}
    \caption{The framework of our proposed method involves initially training a downstream model using real data to identify poorly calibrated bins. Data instances, $x_{i}$, and their prediction probabilities, $\hat{P}(y_{i}|x_{i})$, from these bins are then fed into large language models (LLMs) to generate synthetic data, $x^{syn}_{i}$, along with their corresponding probabilities, $\hat{P}(y_{i}|x^{syn}_{i})$. This synthetic data, combined with the real data, is used to retrain the downstream model, thereby improving calibration outputs without compromising model performance.}
    \label{fig:framework} % Label should be after caption
\end{figure*}

AI risk management is an emerging field that emphasizes understanding the limitations of model predictions. Model calibration techniques are used to address the fact that high accuracy does not always mean high confidence in a model's predictions. For example, consider a classifier trained to recognize handwritten digits. This model might achieve high accuracy on a test set, but it also provides the predicted probability for each class, which reflects its level of uncertainty. If it classifies a digit as a `3' with 70\% probability and as an `8' with 30\% probability, it indicates that while the model predicts `3', it lacks high confidence. Understanding this prediction uncertainty has several key benefits: (1) refining decision-making thresholds to improve overall model performance; (2) adjusting models to perform well under different conditions and data distributions; (3) reducing the black-box nature of machine learning models, fostering greater transparency and trust; and (4) enabling more consistent and reliable decision-making, particularly in risk-sensitive applications where errors can have significant consequences.

Modern deep learning neural networks (NN), however, have been shown to be often miscalibrated i.e., while the NN model performs well in classification, the uncertainty around predictions is also high \cite{wang2021rethinking, minderer2021revisiting}. NLP models trained on classification tasks (such as sentiment analysis) are built on deep learning algorithms, with many hidden layers and regularization steps, and consequently, numerous hyperparameters to be tuned. Recent work has shown the association between increased depth and/or width of NN layers, and miscalibrated outcomes \cite{guo2017calibration}. Model calibration becomes worse in data-scarce scenarios, where the fraction of events predicted does not align with actual outcomes because the amount of data available at hand may not be sufficient enough to be representative across different classes. Data augmentation approaches (e.g., the mixup approach \cite{zhang2017mixup, thulasidasan2019mixup}) and the associated model calibration problems have been discussed in literature \cite{wen2020combining}. But, theories for understanding the association between model performance and calibration are still lacking.

This work is motivated by a recently published paper \cite{sahu-etal-2023-promptmix}, in which LLMs are utilized to generate synthetic data close to the decision boundary to sharpen the discrimination power of the classifier and increase model accuracy. The LLM-generated synthetic data leverage the capability of LLMs in providing both realistic and diverse datasets, which potentially increases the ML model's generalizability on out-of-distribution data. The application of synthetic data has been explored as an augmented training set \cite{van2023synthetic}, validation set \cite{shoshan2023synthetic}, or test set \cite{van2024can} to improve ML model performance. However, applying synthetic data to address model calibration has barely been explored. We aim to use LLM-generated synthetic text data to fine-tune downstream binary classification tasks to reduce expected calibration error (ECE) without sacrificing the ML model's accuracy.

Our approach is derived from the Probably Approximately Correct (PAC) learning framework \cite{valiant1984theory}. We prove that reducing both calibration and misclassification errors can be achieved simultaneously, and we establish the necessity of generating synthetic data for enhancing model calibration. This approach is validated on real-world text datasets. The synthetic data generation process is accomplished using open-source Large Language Models - Llama 2 \cite{touvron2023llama}. Figure \ref{fig:framework} illustrates our proposed framework to improve model calibration and generalization via synthetic data for natural language classification tasks.

The contributions of this work can be summarized as follows:
\begin{enumerate}
    \item We derive the Expected Calibration Error bound to explore the possibility of achieving both high accuracy and low ECE.
    \item We propose a strategy for fixing calibration errors and filling the gaps in the reliability diagram.
    \item We demonstrate the effectiveness of purposefully augmenting LLM-generated synthetic data into the training set to achieve ML model prediction uncertainty calibration.
\end{enumerate}

\section{Calibration Concept}

This section introduces some basic concepts related to model calibration, which would lay a foundation to derive our methodology. 
\\

% \subsection{Calibration Concept}
\noindent\textbf{Expected Calibration Error (ECE)}. ECE is a widely-used metric to evaluate how well a model's predicted probabilities (confidence) align with its actual outcomes (accuracy) \cite{guo2017calibration}. It is calculated by segmenting the full range of predicted probabilities into $M$ equal bins and sorting predictions into these bins based on their confidence. Within each bin, the model's accuracy (the fraction of correct predictions) and average predicted confidence are computed. Let $B_m$ be the set of examples in the $m^{th}$ bin, whose accuracy and confidence are: 
% One notion of miscalibration is the expected difference between
% confidence and accuracy: $\mathbb{E}_{\hat{P}} \left[ \left| P(\hat{Y} = Y \mid \hat{P} = p) - p \right| \right]$. ECE approximates this by binning the predictions in [0, 1] under M equally-spaced intervals, and then taking a weighted average of each bins’ accuracy/confidence difference. Let $B_m$ be the set of examples in the $m^{th}$ bin
% whose predicted confidence falls into interval ($\frac{m-1}{M}, \frac{m}{M}$). The bin $B_m$’s accuracy and confidence are:
\begin{equation}
\label{eq:ac-conf}
\resizebox{0.6\hsize}{!}{$
\begin{aligned}
    \text{Acc}(B_m) &= \frac{1}{|B_m|} \sum_{x_i \in B_m} 1(\hat{y}_i = y_i), \\
    \text{Conf}(B_m) &= \frac{1}{|B_m|} \sum_{x_i \in B_m} \hat{p}_i
\end{aligned}
$}
\end{equation}

where \(1(\hat{y}_i = y_i)\) is an indicator function that is equal to 1 if \(\hat{y}_i = y_i\) and 0 otherwise; \(\hat{p}_i\) is the predicted probability associated with the instance \(x_i\). The concept of accuracy here is based on each class of labels, which is a subset of the well-accepted model evaluation metric: \textit{accuracy}. 
% The confidence for bin \(B_m\) is defined as:
% \[
% \text{Conf}(B_m) = \frac{1}{|B_m|} \sum_{x_i \in B_m} \hat{p}_i,
% \]
% where \(\hat{p}_i\) is the predicted probability associated with the instance \(x_i\). 
The Expected Calibration Error (ECE) given \( n \) examples is defined by taking a weighted average of the absolute differences between the bin's confidence and its accuracy: 
\begin{equation}
\label{eq:ece}
\resizebox{0.75\hsize}{!}{$
\text{ECE} = \sum_{m=1}^M \frac{|B_m|}{n} \left| \text{Acc}(B_m) - \text{Conf}(B_m) \right|$}
\end{equation}

There exist some variants of ECE, like MCE \cite{guo2017calibration}, ACE\cite{nixon2019measuring}, and other metrics to quantify calibration like brier score \cite{rufibach2010use}, however, in this paper, the calibration error refers to ECE.
\\

\noindent\textbf{Reliability Diagram}. Reliability diagram (see Figure \ref{fig:ds1}) is a tool to visualize the model calibration. $Conf(B_{m})$ and $ Acc(B_{m})$ represent the x-axis and y-axis of the diagram respectively for bin $B_{m}$. The diagonal line denotes perfectly calibrated and any deviations from this diagonal line indicate a model's miscalibration. Therefore, the miscalibration can be divided as above the line (underconfidence: $Acc(B_{m})$ > $Conf(B_{m})$) and under the line (overconfidence: $Acc(B_{m})$ < $Conf(B_{m})$) areas. We use a reliability diagram to find out the target bins where synthetic data are needed to fill in.

\section{Methodology}

In this section, we utilize the Probably Approximately Correct (PAC) learning framework to derive the expected calibration error(ECE) bound and discuss the benefits of using synthetic data to improve models' calibration and generalization. Moreover, we use a toy sample to demonstrate our methodology. 

% show the methodology of utilizing synthetic data to improve model calibration and generalizability. We demonstrate our method via both theoretical concepts and a toy example. 

\subsection{From PAC Learning to Expected Calibration Error Bound}
% We have demonstrated that adding synthetic data can improve models’ calibration. But how many synthetic data points do we need given the negligibility of the cost of generating synthetic data? 
%https://engineering.purdue.edu/ChanGroup/ECE595/files/Lecture24_pac.pdf

Probably Approximately Correct (PAC) learning \cite{valiant1984theory} offers a theoretical framework that establishes the bounds on learning model parameters with specified levels of error and confidence, relating model accuracy to confidence level and sample size. According to Hoeffding's inequality,  
\[
\resizebox{0.7\hsize}{!}{
$P(\left| E(h) - E(h^*) \right| > \epsilon ) \leq 2 \exp(-2 \epsilon^2 n)$
}
\]
where $E(h)$ denotes the true error of the hypothesis $h$ on unseen data and $E(h^*)$ denotes the empirical error of the hypothesis $h$ on the training data. Let $\delta$  \footnote{Confidence level in PAC learning refers to a probability that the learned hypothesis with an error rate less than a specified accuracy; confidence in ECE represents the predicted probability for a given prediction.} be the confidence level and make $\delta =  2 \exp(-2 \epsilon_a^2 n)$. This inequality presents the maximum allowable difference between the true and empirical errors based on a given sample size $n$ and desired uncertainty level $\delta$. Thus, we get the minimal sample size to make the hypothesis true considering error difference $\epsilon$ and confidence $\delta$ is given by $n = \log(2/\delta) / (2\epsilon ^2)$. 

Now we derive the ECE bound from the PAC learning framework. First, we extend the
definition of accuracy and confidence in Equation (\ref{eq:ac-conf}) from bin-wise to data-wise. Then we have 
\[\resizebox{0.5\hsize}{!}{$
\begin{aligned}
 \text{Acc}(X) &= \sum_{m=1}^M \frac{|B_m|}{n} \text{Acc}(B_m), \\ \text{Conf}(X) &= \sum_{m=1}^M \frac{|B_m|}{n} \text{Conf}(B_m)
\end{aligned}
$}
\]

% \[
% % \resizebox{0.7\hsize}{!}{$
% \text{Acc}(X) &= \sum_{m=1}^M \frac{|B_m|}{n} \text{Acc}(B_m), \\ \text{Conf}(X) &= \sum_{m=1}^M \frac{|B_m|}{n} \text{Conf}(B_m)
% % $}
% \]

The dataset is denoted by $\{X, y\}_i^n$, where $X$ denotes the feature space, $X \in \{X_{1}, \ldots, X_{n}\}$ and $y$ is the label $y \in \{y_{1}, \ldots, y_{n}\}$. Based on Hoeffding’s inequality, we have
\begin{equation}
\label{eq:lac}
\resizebox{0.8\hsize}{!}{$
P(\left| Acc(X) - Acc(X^*) \right| >\epsilon_a ) \leq 2 \exp(-2 \epsilon_a^2 n)$}
\end{equation}
where $Acc(X)$ denotes the expected accuracy in the model and $Acc(X^*)$ is the observed accuracy of training data. $\epsilon_a$ is the error for accuracy and we let $\delta_a = 2 \exp(-2 \epsilon_a^2 n)$. 
\\
% From equation (1) in section 2, we have: 

% \[
% \text{Acc}(X) = \sum_{m=1}^M \frac{|B_m|}{n} \text{Acc}(B_m),  \quad \text{Conf}(X) = \sum_{m=1}^M \frac{|B_m|}{n} \text{Conf}(B_m)
% \], correspondingly, 
% \[
% \text{Acc}(X^*) = \sum_{m=1}^M \frac{|B_m^*|}{n} \text{Acc}(B_m^*),  \quad \text{Conf}(X^*) = \sum_{m=1}^M \frac{|B_m^*|}{n} \text{Conf}(B_m^*)
% \]

%%%%%%%%%%%%%%%%%%%%%%%%%%%%%%%%%%%%%%%%%%%%%%%%%%%

% \noindent\textbf{Proposition -- Expected Calibration Error Bound.}  \textit{Given $n$ training samples, if the probability of the difference between the expected model parameter and its estimated value being less than $\epsilon_a$ is at least $(1-\delta_a)\%$, then the probability of the difference between the expected calibration error and the estimated calibration error in the training samples being less than $\epsilon_{ECE}$ is at least $(1-\delta_{ECE})\%$. Here, $\delta_{ECE}$ =2$\delta_a$, and $\epsilon_{ECE} =  \epsilon_a + \resizebox{0.4\hsize}{!}{\left|\text{Conf}(X) - \text{Conf}(X^*) \right|} = \epsilon_a + \resizebox{0.6\hsize}{!}{\sum_{m=1}^M \frac{|B_m|}{n}\left| \text{Conf}(B_m) - \text{Conf}(B_m^*)\right|}$.}

\noindent\textbf{Proposition -- Expected Calibration Error Bound.} \textit{Given $n$ training samples, if the probability of the difference between the expected model parameter and its estimated value being less than $\epsilon_a$ is at least $(1-\delta_a)\%$, then the probability of the difference between the expected calibration error and the estimated calibration error in the training samples being less than $\epsilon_{ECE}$ is at least $(1-\delta_{ECE})\%$. Here, $\delta_{ECE} = 2\delta_a$, and $\epsilon_{ECE} = \epsilon_a + $ \resizebox{0.4\hsize}{!}{$\left|\text{Conf}(X) - \text{Conf}(X^*)\right|$} $ = \epsilon_a + $ \resizebox{0.6\hsize}{!}{$\sum_{m=1}^M \frac{|B_m|}{n}\left|\text{Conf}(B_m) - \text{Conf}(B_m^*)\right|$}.}
\\

% $(1-\epsilon_{ECE} =  \epsilon_a +  \resizebox{0.45\hsize}{!}{\left|Conf(X) - Conf(X^*) \right|}$

A shortened proof (detailed proof is provided in Appendix \ref{appendix:b}) is given below: 

By deriving from the left side of equation (\ref{eq:lac}), we get: 
\begin{equation}
\label{eq:ac-eceb}
\resizebox{0.95\hsize}{!}{$
\begin{aligned}
& P\left(\left| \text{Acc}(X) - \text{Acc}(X^*) \right| > \epsilon_a \right) \\
\geq & P\left( \text{ECE}(X)  - \text{ECE}(X^*)  
> \epsilon_a +  \left|\text{Conf}(X) - \text{Conf}(X^*) \right| \right) 
\end{aligned}
$}
\end{equation}

% \begin{flalign*}
% \begin{dmath}
% P\left(\left| \text{Acc}(X) - \text{Acc}(X^*) \right| > \epsilon_a \right) 
% \geq P\left( \text{ECE}(X)  - \text{ECE}(X^*)  > \epsilon_a +  \left|\text{Conf}(X) - \text{Conf}(X^*) \right| \right) 
% \end{dmath}
% \end{flalign*}

Combined with the right side of equation (\ref{eq:lac}): 
\begin{equation}
\resizebox{0.95\hsize}{!}{$
\begin{aligned}
&P\left( \text{ECE}(X)  - \text{ECE}(X^*)  > \epsilon_a + \left|\text{Conf}(X) - \text{Conf}(X^*) \right| \right) \\
&\leq 2 \exp(-2 \epsilon_a^2 n) \\
\\
\Rightarrow & P\left( \left|\text{ECE}(X) - \text{ECE}(X^*) \right| > \epsilon_{\text{ECE}} \right) \\
&\leq 4 \exp(-2 (\epsilon_{\text{ECE}} - \left|\text{Conf}(X) - \text{Conf}(X^*) \right|)^2 n)
\end{aligned}$}
\end{equation}

% \begin{equation}
% \resizebox{0.95\hsize}{!}{$
% \begin{align*}
% % & \begin{aligned}[t] 
% &P\left( \text{ECE}(X)  - \text{ECE}(X^*)  > \epsilon_a + \left|\text{Conf}(X) - \text{Conf}(X^*) \right| \right) \\
% &\leq 2 \exp(-2 \epsilon_a^2 n) 
% % \end{aligned} 
% \\
% \\
% \Rightarrow & \begin{aligned}[t]
% & P\left( \left|\text{ECE}(X) - \text{ECE}(X^*) \right| > \epsilon_{ECE} \right) \\
% &\leq 4 \exp(-2 (\epsilon_{ECE} - \left|\text{Conf}(X) - \text{Conf}(X^*) \right|)^2 n)
% \end{aligned}
% \end{align*}$}
% \end{equation}

% \begin{equation}
% \resizebox{1.0\hsize}{!}{$
% \begin{align*}
% & \begin{aligned}[t] 
% P\left( \text{ECE}(X)  - \text{ECE}(X^*)  > \epsilon_a + \left|\text{Conf}(X) - \text{Conf}(X^*) \right| \right) \\
% &\leq 2 \exp(-2 \epsilon_a^2 n) 
% \end{aligned} \\
% \Rightarrow & \begin{aligned}[t]
% & P\left( \left|\text{ECE}(X) - \text{ECE}(X^*) \right| > \epsilon_{ECE} \right) \\
% &\leq 4 \exp(-2 (\epsilon_{ECE} - \left|\text{Conf}(X) - \text{Conf}(X^*) \right|)^2 n)
% \end{aligned}
% \end{align*}$}
% \end{equation}

% \begin{equation}
% \resizebox{1.0\hsize}{!}{$
% \begin{align*}
% & \begin{aligned}[t] P\left( \text{ECE}(X)  - \text{ECE}(X^*)  >\epsilon_a & +  \left|\text{Conf}(X) - \text{Conf}(X^*) \right|  \right) ) \\
% &\leq 2 \exp(-2 \epsilon_a^2 n) 
% \end{aligned}\\
% \Rightarrow & \begin{aligned}[t]P\left( \left|\text{ECE}(X) & - \text{ECE}(X^*) \right| > \epsilon_{ECE} \right)) \\
% &\leq 4 \exp(-2 (\epsilon_{ECE} - \left|\text{Conf}(X) - \text{Conf}(X^*) \right| )^2 n)
% \end{aligned}
% \end{align*}$}
% \end{equation}

\noindent where $ \epsilon_{\text{ECE}} =  \epsilon_a + \resizebox{0.35\hsize}{!}{$\left|\text{Conf}(X) - \text{Conf}(X^*) \right|$} = \epsilon_a + \resizebox{0.6\hsize}{!}{$\sum_{m=1}^M \left( \frac{|B_m|}{n} \right)\left| \text{Conf}(B_m) - \text{Conf}(B_m^*)\right|$}$. And $n_{\text{ECE}} = \log(4/\delta_{\text{ECE}}) / \left( 2(\epsilon_{\text{ECE}} - \resizebox{0.45\hsize}{!}{$\left| \text{Conf}(X) - \text{Conf}(X^*)\right|$})^2 \right)$.
\\

Hoeffding's inequality holds on Bernoulli random variables and accuracy is computed by counting correct predictions. By introducing the concept of uncertainty $\delta$, we obtain the relationship among error, uncertainty, and sample size for PAC learning. Then we can derive the ECE bound from the same inequality and ECE is a random variable with a value in [0, 1]. Finally, given a sample size, we have the relationship among $\epsilon_a$,$\epsilon_{ECE}$,$\delta_a$, $\delta_{ECE}$. 
\\

\noindent\textit{\textbf{Remark 1:}} \textit{Since the difference between true prediction probabilities and the estimated prediction probabilities exists, given the same data points to train a model, the error for ECE is larger than the error for accuracy compared with the true metrics and the uncertainty level for ECE is two times that for accuracy.}
\\

% \noindent\noindent\textit{\textbf{Remark 2:}} \textit{Based on $n_{ECE} = \log(4/\delta_{ECE}) \\\resizebox{0.65\hsize}{!}{/ (2(\epsilon_{ECE} - \left| \text{Conf}(X) - \text{Conf}(X^*)\right|)^2)}$, increasing the amount of training data will result in smaller error $\epsilon_{ECE}$ and lower uncertainty level  $\delta_{ECE}$; similar to the effects on $\epsilon_{a}$ and $\delta_{a}$.}

\noindent\noindent\textit{\textbf{Remark 2:}} \textit{Based on $n_{ECE} = \log(4/\delta_{ECE})$ \\ \resizebox{0.65\hsize}{!}{$/ (2(\epsilon_{ECE} - \left| \text{Conf}(X) - \text{Conf}(X^*)\right|)^2)$}, increasing the amount of training data will result in smaller error $\epsilon_{ECE}$ and lower uncertainty level $\delta_{ECE}$; similar to the effects on $\epsilon_{a}$ and $\delta_{a}$.}
\\
% \noindent\textit{\textbf{Remark 2:}} \textit{Based on}
% \begin{equation}
% \resizebox{0.8\hsize}{!}{$
% n_{BCE} = \frac{\log(4/\delta_{BCE})}{2(\epsilon_{BCE} - \left| \text{Conf}(X) - \text{Conf}(X^*)\right|)^2}
% $}
% \end{equation}

% \textit{Remark 2} tells us that to increase the amount of training data can both improve model generalizaton and lower expected calibration error. \textcolor
% {blue}{remove: In light of the sparsity of training data in many real world domains, synthetic data offers a promising approach to add more training samples. In addition, synthetic data from out-of-sample distribution can bring diversity and benefit model generalization }\cite{}. \\

% \noindent\textit{\textbf{Remark 3:}} \textit{Becasue of $ \epsilon_{ECE} =  \epsilon_a + \left|\text{Conf}(X) - \text{Conf}(X^*) \right|$, to reduce $\left|\text{Conf}(X) - \text{Conf}(X^*) \right|$ can make error for ECE align with the error for accuracy, making models achieve good calibration and better generalization at the same time.}

% \noindent\noindent\textit{\textbf{Remark 3:}} \textit{Since $\epsilon_{ECE} = \epsilon_a + \resizebox{0.39\hsize}{!}{\left|\text{Conf}(X) - \text{Conf}(X^*)\right|}$, reducing \resizebox{0.4\hsize}{!}{$\left|\text{Conf}(X) - \text{Conf}(X^*)\right|$} aligns the ECE error with the accuracy error. This alignment helps models achieve both good calibration and better generalization.}

\noindent\noindent\textit{\textbf{Remark 3:}} \textit{Since $\epsilon_{ECE} = \epsilon_a + \resizebox{0.39\hsize}{!}{$\left|\text{Conf}(X) - \text{Conf}(X^*)\right|$}$, reducing \resizebox{0.4\hsize}{!}{$\left|\text{Conf}(X) - \text{Conf}(X^*)\right|$} aligns the ECE error with the accuracy error. This alignment helps models achieve both good calibration and better generalization.}

% \noindent\textit{\textbf{Remark \hspace{-0.4cm}3:}} \textit{since the equation $\epsilon_{ECE} = \epsilon_a + \resizebox{0.4\hsize}{!}{\left|\text{Conf}(X) - \text{Conf}(X^*)\right|}$, reducing \resizebox{0.4\hsize}{!}{$\left|\text{Conf}(X) - \text{Conf}(X^*)\right|$} aligns the ECE error with the accuracy error. This alignment helps models achieve both good calibration and better generalization.}
% \\

% \textbf{Proposition 4} \textit{if minimal data for parameters learnable for accuracy is $n_a = \log(2/\delta_a) / (2 \epsilon_a ^2)$, then the minimal data for parameters learnable for calibration error is $n_{BCE} = \log(4/\delta_{BCE}) / (2(\epsilon_{BCE} - \left| \text{Conf}(X) - \text{Conf}(X^*)\right|)^2)$}

\textit{Remark 1} explains why some neural networks own a good performance but are more likely to be ill-calibrated. \textit{Remark 2} indicates that increasing the amount of training data can both improve model generalization and lower expected calibration error. When the training data is insufficient, synthetic data is the natural option to augment the training data size. \textit{Remark 3} provides insights into what kind of synthetic data is needed to fix the calibration issue. The newly added synthetic data should reduce the difference between predicted probabilities and the true probabilities. We cannot know the true difference of $\left|\text{Conf}(X) - \text{Conf}(X^*) \right|$, but we can decrease it by reducing $\left|\text{Conf}(B_{m}) - \text{Conf}(B_{m}^*) \right|$ in bins that display the gaps given that $\text{Conf}(X) = \sum_{m=1}^M \frac{|B_m|}{n} \text{Conf}(B_m)$, as we know where the perfect calibration line is for each bin. Therefore, we can manipulate the prediction probability of synthetic data to minimize the difference. In other words, synthetic data is applied to fill the gaps against the perfect calibration. That is, we try to decrease ECE by using synthetic data to lower the ECE bound. 

% \begin{figure*}[ht]
% \begin{subfigure}{0.42\textwidth}
%   \includegraphics[width=\linewidth]{figures/ds1.final.png}
%   \caption{Reliability diagram with the target bin}
%   \label{fig:ds1}
% \end{subfigure}
% \hspace{1cm}
% \begin{subfigure}{0.5\textwidth}
%   \includegraphics[width=1.02\linewidth, height=4.98cm]{figures/ds.final.png}
%   \caption{Move generated synthetic data away from DB}
%   \label{fig:ds2}
% \end{subfigure} 
% \label{fig:db}
% \caption{Generating synthetic data to address miscalibration gaps. In (a), the target bin for calibration is identified as \textit{Low Probability \& Overconfidence}. Synthetic data is generated away from the decision boundary in (b).}
% \end{figure*}

\begin{figure*}[ht]
    \centering % Center the figure
    \begin{subfigure}{0.42\textwidth}
        \begin{adjustbox}{width=\linewidth} % Adjust box to the width of the subfigure
            \includegraphics{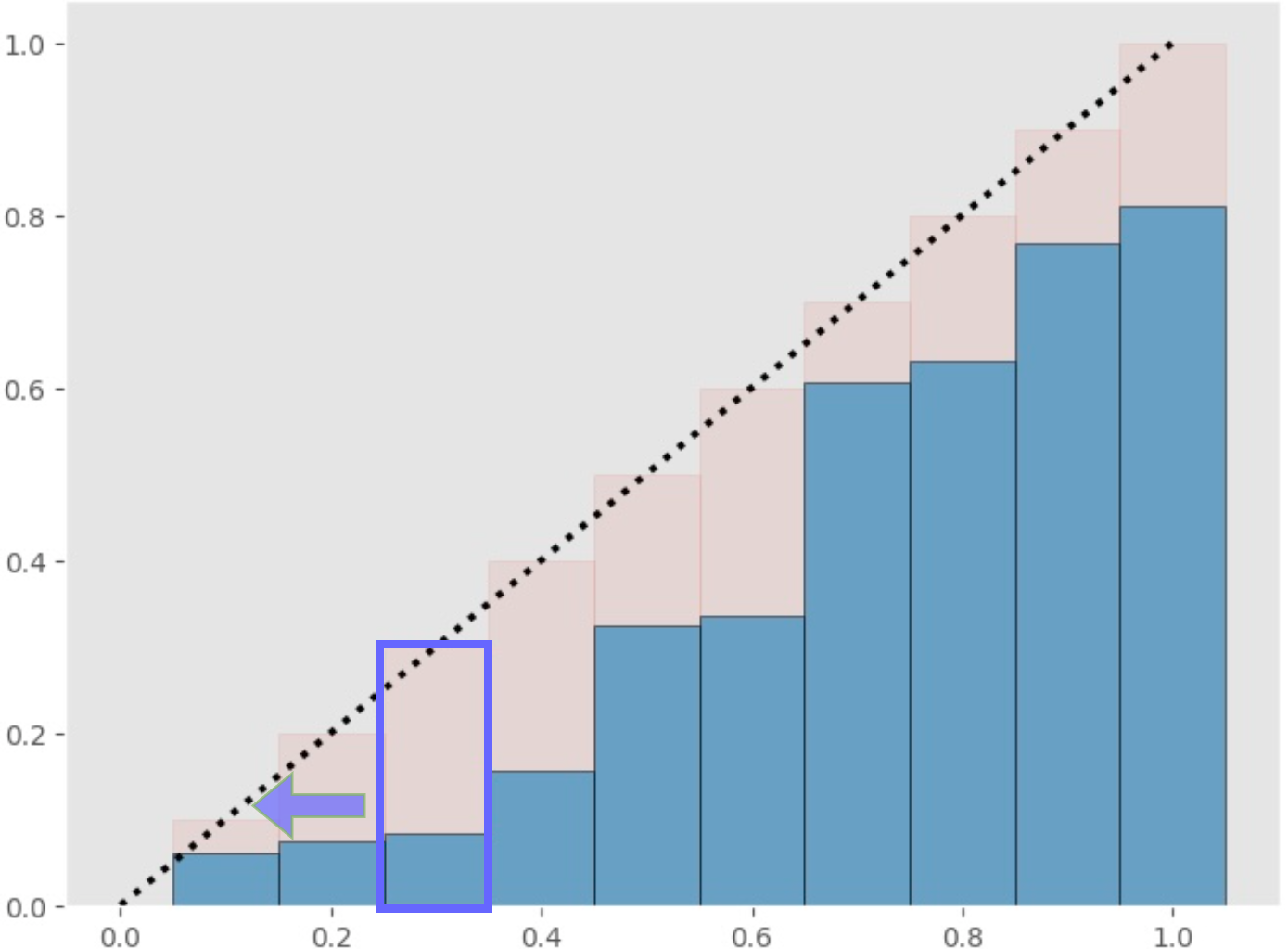}
        \end{adjustbox}
        \caption{Reliability diagram with the target bin}
        \label{fig:ds1}
    \end{subfigure}
    \hspace{1cm} % Space between subfigures
    \begin{subfigure}{0.5\textwidth}
        \begin{adjustbox}{width=\linewidth, height=4.98cm} % Adjust box to maintain height
            \includegraphics{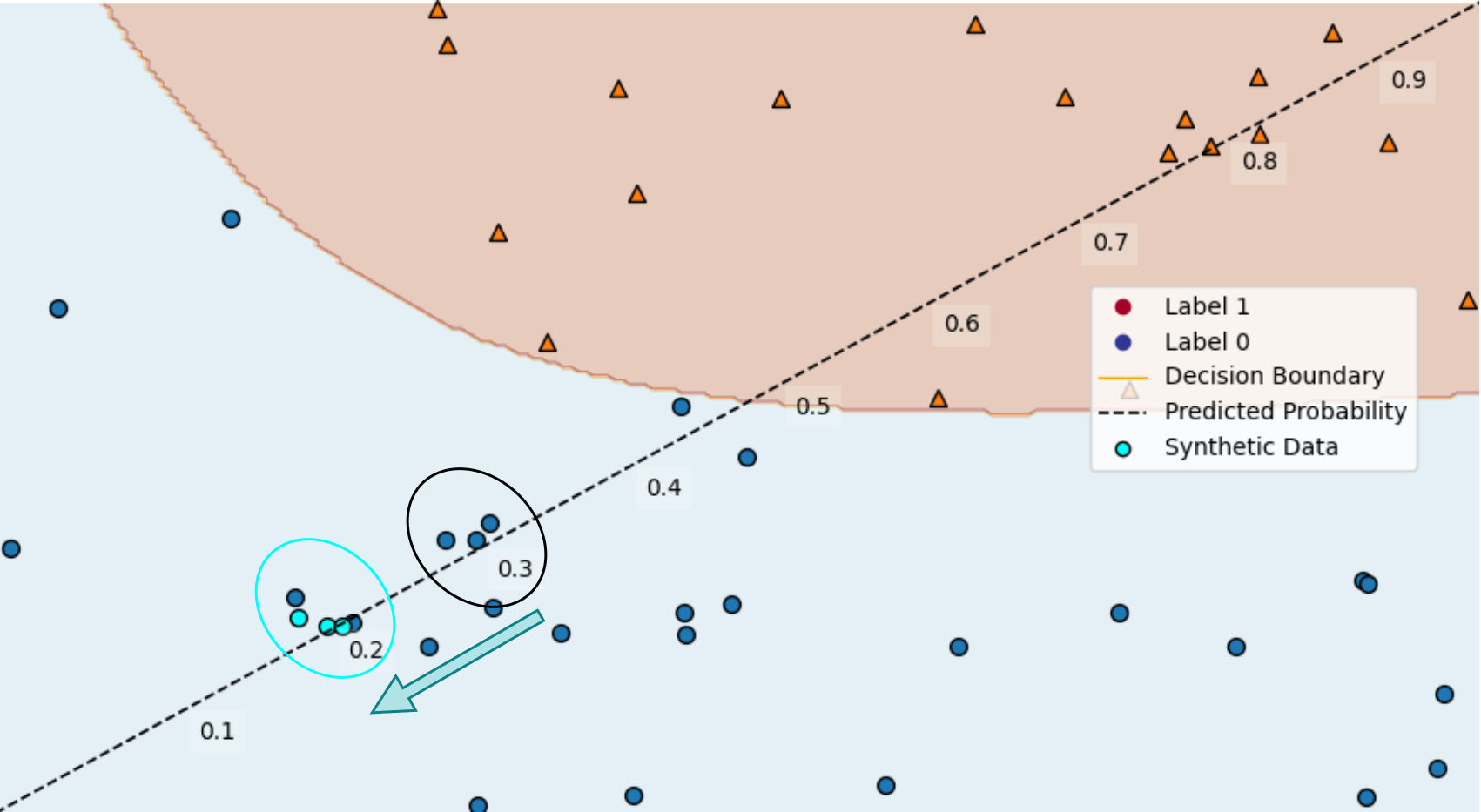}
        \end{adjustbox}
        \caption{Move generated synthetic data away from DB}
        \label{fig:ds2}
    \end{subfigure}
    \caption{Generating synthetic data to address miscalibration gaps. In (a), the target bin for calibration is identified as \textit{Low Probability \& Overconfidence}. Synthetic data is generated away from the decision boundary in (b).}
    \label{fig:db} % Label after caption
\end{figure*}

\subsection{Synthetic Data Generation Strategy}

% \begin{figure*}[ht]
% \begin{subfigure}{0.42\textwidth}
%   \includegraphics[width=\linewidth]{figures/ds1.final.png}
%   \caption{Reliability diagram with the target bin}
%   \label{fig:ds1}
% \end{subfigure}
% \hspace{1cm}
% \begin{subfigure}{0.5\textwidth}
%   \includegraphics[width=1.02\linewidth, height=4.98cm]{figures/ds.final.png}
%   \caption{Move generated synthetic data away from DB}
%   \label{fig:ds2}
% \end{subfigure} 
% \label{fig:db}
% \caption{Generating synthetic data to address miscalibration gaps. In (a), the target bin for calibration is identified as \textit{Low Probability \& Overconfidence}. Synthetic data is generated away from the decision boundary in (b).}
% \end{figure*}

Synthetic data generation consists of two stages: First, we specify the gaps against the perfect calibration line in the reliability diagram. Bins over the line are underconfident while those under the line are overconfident. The data points in those bins are the target data samples for synthetic data generation. With the predicted probability 0.5 as the cutoff, we categorize the reliability diagram (Figure \ref{fig:ds1}) into four scenarios: \textit{Low Probability \& Over Confidence, Low Probability \& Under Confidence, High Probability \& Over Confidence, and High Probability \& Under Confidence}, as shown in Table \ref{table-synGen}.

% \begin{table*}[h]
%   \label{table-synGen}
%   \centering
%   \begin{tabular}{lll}
%     \toprule
%       & Over Confidence   & Under Confidence \\
%     \midrule
%      \thead{Low Probability 
%      \\($\hat{P}(y_{i}|x_{i})$<=0.5)}  & \thead{Decrease predicted prob \\ (Move away DS)}  &  \thead{Increase predicted prob \\(Move towards DS)}      \\
%     \thead{ High Probability \\
%      ($\hat{P}(y_{i}|x_{i})$>0.5)}& \thead{Increase predicted prob \\(Move towards DS)}  &  \thead{Increase predicted prob \\(Move away DS)}   \\
%     \bottomrule
%   \end{tabular}
%   \caption{Synthetic Data Generation Method (DS: Decision boundary)}
% \end{table*}

% \begin{table}[t]
%   \centering
%   %\small
%   \scriptsize
%   \setlength{\tabcolsep}{12pt} % Adjust the column separation
%   \begin{tabular}{@{}p{1cm}cc@{}}
%     \toprule
%     & \textbf{Over Confidence} & \textbf{Under Confidence} \\
%     \midrule
%     \textbf{\resizebox{!}{12pt}{\thead{Low Probability \\ ($\hat{P}(y_{i}|x_{i}) \leq 0.5$)}}}
%     & \resizebox{!}{12pt}{\thead{Decrease predicted prob \\ (Move away from DB)}}
%     & \resizebox{!}{12pt}{\thead{Increase predicted prob \\ (Move towards DB)}} \\
%     \textbf{\resizebox{!}{12pt}{\thead{High Probability \\ ($\hat{P}(y_{i}|x_{i}) > 0.5$)}}}
%     & \resizebox{!}{12pt}{\thead{Increase predicted prob \\ (Move towards DB)}}
%     & \resizebox{!}{12pt}{\thead{Increase predicted prob \\ (Move away from DB)}} \\
%     \bottomrule
%   \end{tabular}
%   \caption{Synthetic Data Generation Strategy (DB: Decision Boundary)}
%   \label{table-synGen}
% \end{table}

Next, LLMs, which serve as text generators, are used to create synthetic text data. Since LLMs are trained on diverse and extensive data spanning a wide range of sources, we can distill the knowledge from LLMs to generate synthetic data that is considered out-of-distribution of training data. We ask LLMs to imitate the classifier we trained by generating similar instances using data samples we collected from the target bin. Specifically, we pass the data instance $x_i$ and $\hat{P}(y_{i}|x_{i})$ from a trained classifier to LLMs and ask it to generate a similar instance $x^{syn}_{i}$ with $\hat{P}(y_{i}|x^{syn}_{i})$, where $|\hat{P}(y_{i}|x_{i})$ - $\hat{P}(y_{i}|x^{syn}_{i})|$ = $\left|\text{Conf}(B_{m}) - \text{Conf}(B_{m}^*) \right|$. For example, suppose there are $n_{bins}$ bins, if the target text is from $m^{th}$ bin and $\left|\text{Conf}(B_m) - \text{Conf}(B_m^*) \right|$ is $\alpha$, then we will ask LLMs to generate the synthetic texts that have the $\frac{m}{n_{bins}} \pm \alpha$ probability belonging to one class and the $1- (\frac{m}{n_{bins}} \pm \alpha)$ probability for the other class. 

\begin{table}[h]
  \centering
  %\small
  \scriptsize
  \setlength{\tabcolsep}{12pt} % Adjust the column separation
  \begin{tabular}{@{}p{1cm}cc@{}}
    \toprule
    & \textbf{Over Confidence} & \textbf{Under Confidence} \\
    \midrule
    \textbf{\resizebox{!}{12pt}{\thead{Low Probability \\ ($\hat{P}(y_{i}|x_{i}) \leq 0.5$)}}}
    & \resizebox{!}{12pt}{\thead{Decrease predicted prob \\ (Move away from DB)}}
    & \resizebox{!}{12pt}{\thead{Increase predicted prob \\ (Move towards DB)}} \\
    \textbf{\resizebox{!}{12pt}{\thead{High Probability \\ ($\hat{P}(y_{i}|x_{i}) > 0.5$)}}}
    & \resizebox{!}{12pt}{\thead{Increase predicted prob \\ (Move towards DB)}}
    & \resizebox{!}{12pt}{\thead{Increase predicted prob \\ (Move away from DB)}} \\
    \bottomrule
  \end{tabular}
  \caption{Synthetic Data Generation Strategy (DB: Decision Boundary). Refer to Appendix \ref{appendix:a} for the prompts used for data generation across different scenarios.}
  \label{table-synGen}
\end{table}

To illustrate our method, in Figure \ref{fig:ds2} a hidden predicted probability line is shown orthogonal to the estimated decision boundary. Data points close to this decision boundary would be predicted with around 0.5 probability (the softmax output), while data points at the two ends of this line could be predicted with close to 0.1 or 0.9 probability, respectively. Suppose that our targeted bin has a confidence of 0.3 and it is over-confident as shown in Figure \ref{fig:ds1} (highlighted in a purple rectangular). The gap between the empirical and theoretical uncertainty values is shown in red. There are two possible solutions to fill the gap: 1) increasing the number of incorrect predictions, thus raising the blue bar that represents the empirical inaccurate prediction percentage; or 2) moving this bin to the left, into the bin with a smaller uncertainty value. We use the second solution to align the miscalibration bins because the first solution could harm the accuracy of the classifier. % Moving the decision boundary too much by adding similar incorrect instances could degrade its performance.

\begin{figure*}[!h]
    \centering
    % First figure
    \begin{subfigure}{0.24\textwidth}
        \centering
        \begin{adjustbox}{width=\linewidth} % Adjust box to the width of the subfigure
            \includegraphics{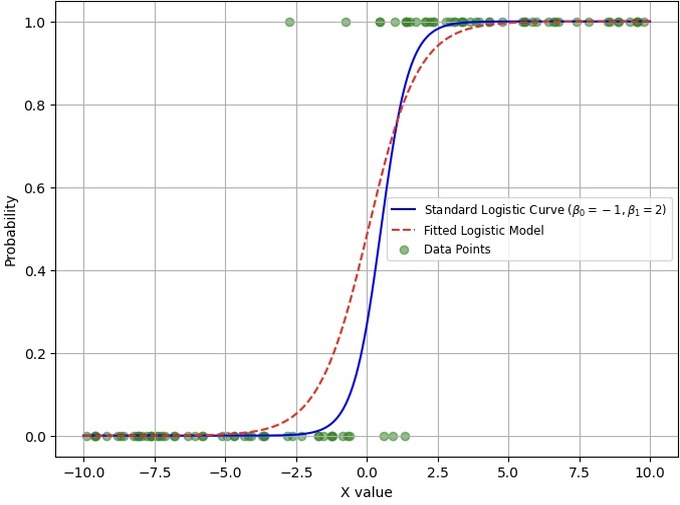}
        \end{adjustbox}
        \caption{\label{fig:lr}}
    \end{subfigure}
    \hfill
    % Second figure
    \begin{subfigure}{0.24\textwidth}
        \centering
        \begin{adjustbox}{width=\linewidth} % Adjust box to the width of the subfigure
            \includegraphics{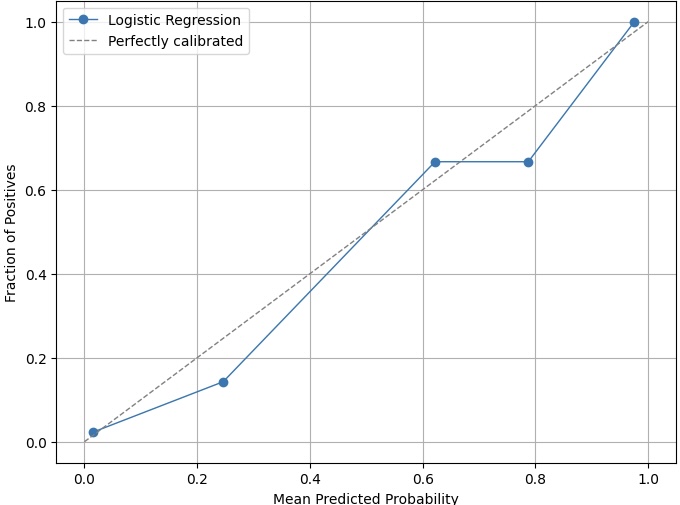}
        \end{adjustbox}
        \caption{\label{fig:rd}}
    \end{subfigure}
    \hfill
    % Third figure
    \begin{subfigure}{0.24\textwidth}
        \centering
        \begin{adjustbox}{width=\linewidth} % Adjust box to the width of the subfigure
            \includegraphics{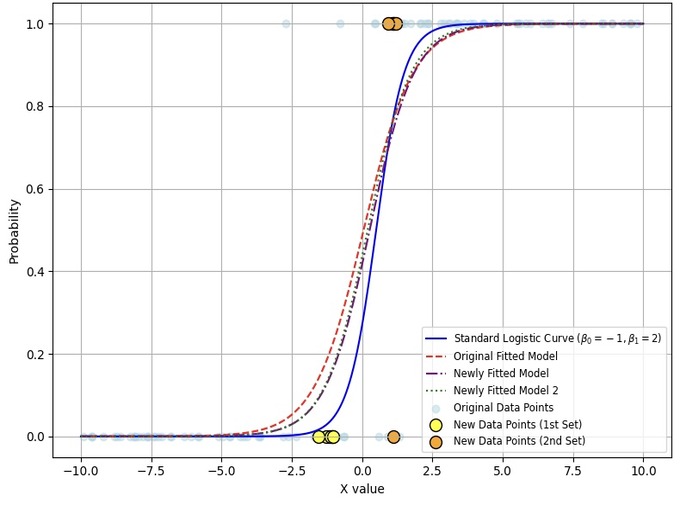}
        \end{adjustbox}
        \caption{\label{fig:synCurve}}
    \end{subfigure}
    \hfill
    % Fourth figure
    \begin{subfigure}{0.24\textwidth}
        \centering
        \begin{adjustbox}{width=\linewidth} % Adjust box to the width of the subfigure
            \includegraphics{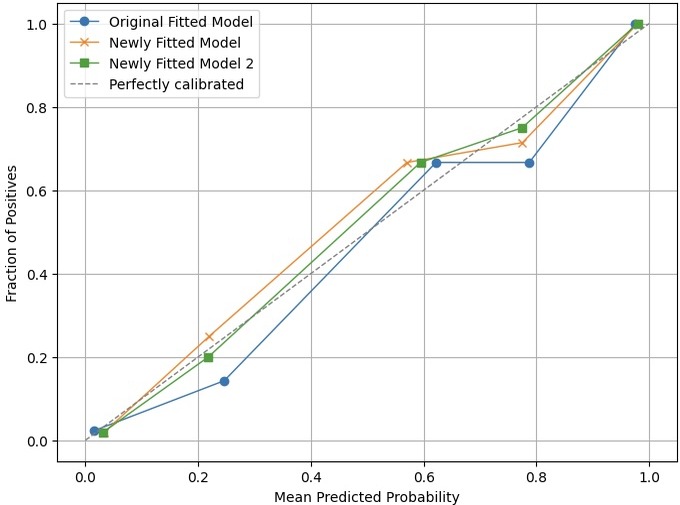}
        \end{adjustbox}
        \caption{\label{fig:RGsyn}}
    \end{subfigure}
    
    \caption{The iterative process of enhancing the accuracy and calibration of a 1D logistic regression model is demonstrated. Initially, the model is fitted using observed data (a), followed by the creation of its reliability diagram to identify poorly calibrated bins (b). Next, synthetic data points are strategically added to two targeted bins and the model is refitted. This iterative approach results in the model closely approximating the true logistic curve (c), thereby improving the calibration in the reliability diagram (d).}
    \label{fig:toy}
\end{figure*}

In Figure \ref{fig:ds2}, the target data points are circled in black and the synthetic data are in the blue circle, which are generated based on the generation strategy in Table \ref{table-synGen}. Since the synthetic data shares a similar feature space and the same labels as the target data samples, the retrained classifier would predict them as the same class but with smaller probabilities. This makes it more likely for the synthetic data to be assigned to the same bin as the original target data samples. In this way, we push the predicted probability of this bin away from the decision boundary and reduce the difference $\left|\text{Conf}(B_{m}) - \text{Conf}(B_{m}^*) \right|$.

Utilizing LLM, we employ a two-stage approach to ensure both the fidelity and diversity of the synthetic data generated. We obtain the instance $x^{syn}_{i}$ with probability $\hat{P}(y^{syn}_{i}|x^{syn}_{i})$ in the first stage and relabel it via LLMs to ensure it belongs to the same class of $x_i$ in the second stage. Since step 1 mixes up information from two labels to some degree, it enhances data diversity. The relabeling of the second stage confirms that the generated texts belong to the correct label, which guarantees its fidelity. 

\subsection{Toy Example}

% \begin{figure*}[t]
%   \includegraphics[width=0.24\linewidth]{figures/c1.jpg} \hfill
%   \includegraphics[width=0.24\linewidth]{figures/RD.jpg} \hfill
%   \includegraphics[width=0.24\linewidth]{figures/synCurve 2.jpg} \hfill
%   \includegraphics[width=0.24\linewidth]{figures/RGsyn.jpg}
%   \caption {A toy example}
%   \label{fig:lr}
% \end{figure*}

We use a 1D logistic regression classifier as an example to demonstrate that adding appropriate synthetic data in the target bins can produce a better-calibrated and more accurate model. Parameters of the true model are defined: $\beta_0$ = -1 and $\beta_1$ = 2. We randomly simulate 300 data points from the range between -10 and 10 and classify them based on the true model as the label. A logistic regression model is fitted on these data points. The fitted parameters are $\beta_0$ = -0.06 and $\beta_1$ = 1.13. The model achieves an accuracy of 0.95 and an ECE of 0.0405 (Figure \ref{fig:lr}).

Figure \ref{fig:rd} shows us that the fitted logistic regression is overconfident about its predictions in the $2^{nd}$ bin and $4^{th}$ bin. Now we target these two bins to generate some synthetic data points to fill the gap. The function we used to generate synthetic data points is a left-sided truncated normal distribution, whose parameters are: $\mu$ = $\mu_{Bin_i}$,  $SD$ = $SD_{Bin_i}$, n = $|Bin_i|$, $i = \{2, 4\}$.  We add new data points step by step to see how the logistic curve changes: 1) add synthetic data of the $2^{nd}$ bin, 2) then add synthetic data of the $4^{th}$ bin based on previously added data points. See the parameters and performance for newly fitted models below: 

\begin{itemize}
    \item orginal data (\textbf{orginal fitted model}): $\beta_0$ = -0.06 and $\beta_1$ = 1.13, ACC: 0.95, ECE: 0.0405; 
    \item synthetic data in $2^{nd}$ bin (\textbf{newly fitted model}): $\beta_0$ = -0.339 and $\beta_1$ = 1.2627, ACC: 0.95327, ECE: 0.0424; 
    \item synthetic data in $2^{nd}$ bin + $4^{th}$ bin (\textbf{newly fitted model 2}): $\beta_0$ = -0.2558 and $\beta_1$ = 1.2953, ACC: 0.9469, ECE: 0.0366.
\end{itemize}

% target bin2 and bin5. add bin2 first and bin5 second. 
% bin2: -0.3390929807221772, 1.2627080998990687, ACC: 0.9532710280373832, ECE: 0.0424. 
% bin2 + bin5: beta0: -0.2558015936769575, beta1: 1.2952545923397398, ACC: 0.9469026548672567, 
% ECE: 0.0366. 

% \begin{figure}
% \caption{\label{fig:synCurve} Fitted logistic regression on the original data and synthetic data}
% \centering
% \includegraphics[width=0.8\textwidth]{figures/synCurve.png}
% \end{figure}

% \begin{figure}
% \caption{\label{fig:RGsyn} Reliability diagram on the original data and synthetic data}
% \centering
% \includegraphics[width=0.8\textwidth]{figures/RGsyn.png}
% \end{figure}

Figure \ref{fig:synCurve} and Figure \ref{fig:RGsyn} illustrate that incorporating synthetic data generated from overconfident bins shifts the fitted logistic curve towards the standard logistic curve, resulting in a more accurately calibrated model.

\setcounter{table}{1}
\begin{table}[ht]
  \centering
  \footnotesize	 % Reduced font size
  \setlength{\tabcolsep}{4pt} % Adjust horizontal space between columns
  \begin{tabular}{@{}lcccc@{}}
    \toprule
    \textbf{Dataset} & \textbf{Classes} & \textbf{Balanced} & \textbf{Class Prop.} & \textbf{\#Train/\#Test} \\
    \midrule
    \textbf{TC}   & 2 & No   & 65:35 & 3104/345 \\
    \textbf{SUBJ} & 2 & Yes  & 50:50 & 8000/2000 \\
    \textbf{B77}  & 2 & Yes  & 44:56 & 177/80 \\
    \textbf{SE}   & 2 & No   & 35:65 & 3012/750 \\
    \textbf{Arxiv}  & 2 & Yes  & 50:50 & 4800/1200 \\
    \textbf{Medical}   & 2 & Yes   & 50:50 & 2662/1060 \\
    \bottomrule
  \end{tabular}
  \caption{Dataset Statistics}
  \label{table-dataset}
\end{table}

\section{Experiment}

\subsection{Dataset}

\setcounter{table}{2}
\begin{table*}[!h]
    \centering
    \small
    \begin{adjustbox}{width=\textwidth} % Adjust the table to fit the text width
    \begin{tabular}{|c |p{1.6cm} |p{3cm} |p{4cm}| p{3cm}| p{1cm} |}
        \hline
        \textbf{Dataset} & \multicolumn{2}{c|}{\textbf{Input}} & \centering \textbf{Prompt} & \textbf{\thead{Generated Text\\ ($x^{syn}_{i}$ , $\hat{P}(y_{i}|x^{syn}_{i})$)}} & \textbf{\thead{Label \\(Y)}} \\ \hline
        \multirow{11}{*}{\textbf{SE}} & \centering  \multirow{9}{*}{$x_{i}$} & \vspace{0.005mm} The zoom function on this camera is so loud that sometimes you will be unable to use it if you find yourself in a situation where you must be quiet. & \vspace{1.5mm} \multirow{2}{4cm}{An example $x_i$ which belongs 75\% to negative and 25\% to positive (based on a classifier's categorization). Now I ask you to act as that classifier and based on this example, generate a diverse set of 3 short utterances where each utterance belongs 55\% to negative and 45\% to positive.}  & \vspace{4mm} \multirow{5}{3cm}{I love how this router can handle a large network, but the price is a bit steep for my taste. (55\% negative, 45\% positive)} &  \multirow{11}{*}{negative}\\ \cline{2-3}
        
        & \centering \multirow{2}{*}{$\hat{P}(y_{i}|x_{i})$ }& \centering \multirow{2.5}{*}{ 0.75} \newline & &  &  \\ \cline{2-3}
                     
        & \centering miscalibration type &  \centering \multirow{2}{*}{ overconfidence}  \newline & & & \\ \hline
    \end{tabular}
    \end{adjustbox}
    \caption{An example of generating synthetic data via LLM. As an example, we use the SE dataset; for information on other datasets, see Table \ref{tab:prompt} in Appendix \ref{appendix:d}. \textbf{Input} contains the original text ($x_{i}$) and the average predictive probability of the bin it comes from ($\hat{P}(y_{i}|x_{i})$). \textbf{Generated Text} is the one after the relabeling process. \textbf{Note}: during re-fine tuning of the downstream model, we exclude $\hat{P}(y_{i}|x^{syn}_{i})$ -- (55\% negative, 45\% positive) and retain only $x^{syn}_{i}$ in the dataset.}
    \label{tab:prompt-example}
\end{table*}

We employ four datasets on text classification tasks across multiple domains with varying sample sizes and proportions of class. To better evaluate our approach, we select two balanced and two imbalanced datasets, respectively, and the sample size varies from hundreds to thousands. The Complaints dataset (TC) \cite{preotiuc2019automatically} contains 3K tweets regarding product reviews, which are categorized as complaints or not-a-complaints. The SUBJectivity dataset (SUBJ) \cite{pang2004sentimental} is a benchmark dataset that contains 10K objective/subjective movie reviews. Banking77 (B77) \cite{casanueva2020efficient} is a dataset comprising fine-grained intents within the banking domain featuring multiple classes. For our study, we select instances from two of these classes, which makes the size of the dataset relatively small. SentEval (SE) \cite{hu2004mining} contains 3K data used for sentiment analysis tasks.

Additionally, to discover whether the pre-trained knowledge of LLMs is a crucial element in determining the performance of generated synthetic data on downstream tasks. We choose two newly-released datasets that are unlikely to be a part of the training data of LLMs we used in the paper (\textit{Llama-2}): Arxiv-10 \cite{farhangi2022protoformer} and Medical \cite{fansi2022ddxplus}. We pick ‘cs’ and ‘stat’ classes, and randomly sample 30\% from the entire data in Arxiv-10. Its task is to classify the subject based on the title of a paper. Medical is the dataset in a medical diagnosis domain categorizing a specific disease based on a patient’s symptoms, where "Influenza" and "Anaphylaxis" in our experiment. See detailed statistics of these datasets in Table \ref{table-dataset}.

\subsection{Training}
All datasets are split into training, validation, and test sets. For the TC dataset, we split the entire data into training, validation, and test sets with a ratio of 80:10:10. For other datasets, the validation sets are created by randomly sampling 20\% from the training set. All experiments are evaluated on their original test sets. We fine-tune BERT$_{base}$ \cite{devlin2018bert} models for text classification by adding a dropout and softmax layer following the pre-trained structure. We train each model 5 epochs and apply a 1e-6 learning rate and 50\% dropout. 

% \setcounter{table}{1}
% \begin{table}[ht]
%   \centering
%   \small
%   \begin{tabular}{llllll}
%     \toprule
%       & \textbf{TC}   & \textbf{SUBJ}     & \textbf{B77}  & \textbf{SE}\\
%     \midrule
%     Classes & 2 & 2 & 2 & 2 \\
%     Balanced & No   & Yes  & Yes  & No   \\
%     Class Proportion &  65:35 & 50:50  &  44:56& 35:65\\
%     \# Train  & 3104 & 8000  &  177   & 3012\\
%     % \# Validation & 345     &    &   &    \\
%     \# Test   & 345 & 2000    & 80  & 750\\
%     \bottomrule
%   \end{tabular}
%   \caption{Dataset Stastics}
%   \label{table-dataset}
% \end{table}

% \setcounter{table}{1}
% \begin{table}[ht]
%   \centering
%   \footnotesize	 % Reduced font size
%   \setlength{\tabcolsep}{4pt} % Adjust horizontal space between columns
%   \begin{tabular}{@{}lcccc@{}}
%     \toprule
%     \textbf{Dataset} & \textbf{Classes} & \textbf{Balanced} & \textbf{Class Prop.} & \textbf{\#Train/\#Test} \\
%     \midrule
%     \textbf{TC}   & 2 & No   & 65:35 & 3104/345 \\
%     \textbf{SUBJ} & 2 & Yes  & 50:50 & 8000/2000 \\
%     \textbf{B77}  & 2 & Yes  & 44:56 & 177/80 \\
%     \textbf{SE}   & 2 & No   & 35:65 & 3012/750 \\
%     \bottomrule
%   \end{tabular}
%   \caption{Dataset Statistics}
%   \label{table-dataset}
% \end{table}

\textbf{Step 1.} After completing the training process, we calculate the reliability diagram and the difference ($D$) between the proportion of positive labels and the mean predicted values for each bin based on the validation set. If the absolute value of $D_m$ in $m^{th}$ bin is larger than the threshold 0.03, the data in the $m^{th}$ bin will be selected to generate synthetic data. 

\textbf{Step 2.} We use LLMs to generate synthetic data based on the texts in the target bin from Step 1. To explore the effect of the number of bins (M), we select three scenarios by setting M = 10, 15, 20.

\setcounter{table}{3}
\begin{table*}[t]
  \centering
  \scriptsize
  \setlength{\tabcolsep}{4pt} % Reducing space between columns
  \begin{tabularx}{\textwidth}{l|*{14}{>{\centering\arraybackslash}X}} % Updated for 14 columns
    \toprule
    & \multicolumn{2}{c|}{\textbf{TC}} & \multicolumn{2}{c|}{\textbf{SUBJ}} & \multicolumn{2}{c|}{\textbf{B77}} & \multicolumn{2}{c|}{\textbf{SE}} & \multicolumn{2}{c|}{\textbf{Arxiv}} & \multicolumn{2}{c}{\textbf{Medical}} \\ % Adjusted column headers
    \cmidrule(r){2-13}
    \textbf{Metric}  & ACC & ECE  & ACC & ECE   & ACC & ECE  & ACC & ECE & ACC & ECE & ACC & ECE \\ % Updated row headers
    \midrule

    Baseline  & \makecell{0.867 \\ (0.00)} & \makecell{0.058 \\ (0.02)}  & \makecell{0.955 \\ (0.01)} & \makecell{0.034 \\ (0.01)}  & \makecell{0.708 \\ (0.12)} & \makecell{0.234 \\ (0.04)}  & \makecell{0.884 \\ (0.01)} & \makecell{0.06 \\ (0.00)} & \makecell{0.805 \\ (0.00)} & \makecell{0.105 \\ (0.01)} & \makecell{0.864  \\ (0.00)} & \makecell{0.051 \\  (0.01)} \\ % Replace "X" with actual values
    Isotonic & \makecell{0.871 \\ (0.00)} & \makecell{0.082 \\ (0.01)} & \makecell{0.959 \\ (0.00)} & \makecell{0.027 \\ (0.01)} & \makecell{0.850 \\ (0.02)} &  \makecell{0.063 \\ (0.01)} & \makecell{0.890 \\ (0.01)}  & \makecell{0.058 \\ (0.01)} & \makecell{0.812 \\ (0.01)} & \makecell{0.114 \\  (0.01)} & \makecell{0.869  \\  (0.01)} & \makecell{0.069  \\ (0.01)} \\
    Platt scaling & \makecell{0.863 \\ (0.01)} & \makecell{0.086 \\ (0.01)} & \makecell{0.955 \\ (0.01)} & \makecell{0.029 \\ (0.00)} & \makecell{0.846 \\ (0.03)} & \makecell{0.207 \\ (0.03)} & \makecell{0.888 \\ (0.01)} & \makecell{0.068 \\ (0.00)} & \makecell{0.807 \\ (0.01)} & \makecell{0.122 \\ (0.00)} & \makecell{0.869  \\ (0.01)} & \makecell{0.065 \\  (0.01)}  \\
    MC dropout & \makecell{0.868 \\ (0.02)} & \makecell{0.054 \\ (0.01)} & \makecell{0.952 \\ (0.01)} & \makecell{0.032 \\ (0.01)} &  \makecell{0.821 \\ (0.23)} & \makecell{0.274 \\ (0.14)}  & \makecell{0.876 \\ (0.01)} &  \makecell{0.050 \\ (0.02)} & \makecell{0.799\\ (0.01)} & \makecell{ 0.058 \\ (0.04)} & \makecell{0.871 \\ (0.01)} & \makecell{0.070 \\ (0.01)}  \\
    Temp scaling & \makecell{0.867 \\ (0.01)} & \makecell{0.049 \\ (0.01)} & \makecell{0.955 \\ (0.01)} & \makecell{0.026 \\ (0.01)}  & \makecell{0.708 \\ (0.12)}  & \makecell{0.253 \\ (0.17)} & \makecell{0.884 \\ (0.01)} & \makecell{0.038 \\ (0.00)} & \makecell{0.805 \\ (0.00)} & \makecell{0.070 \\ (0.01)} & \makecell{0.864\\ (0.00)} & \makecell{ 0.056  \\ (0.01)}   \\
    \midrule
    \textbf{10 bins} \\
    \midrule

    Synthesis & \makecell{0.867 \\ (0.01)} & \makecell{0.053 \\ (0.01)} & \makecell{0.960 \\ (0.01)} & \makecell{0.027 \\ (0.01)} & \makecell{0.625 \\ (0.07)} & \makecell{0.255 \\ (0.10)} & \makecell{0.871 \\ (0.00)} & \makecell{0.055 \\ (0.02)} & \makecell{0.815\\  (0.01)} & \makecell{0.077 \\  (0.03)} & \makecell{0.873 \\ (0.01)} & \makecell{0.048 \\ (0.01)} \\
    Synthesis+ & \makecell{0.886 \\ (0.01)} & \makecell{0.046 \\ (0.01)} & \makecell{0.961 \\ (0.00)} & \makecell{0.03 \\ (0.00)} & \makecell{0.792 \\ (0.20)} & \makecell{0.231 \\ (0.03)} & \makecell{0.889 \\ (0.01)} & \makecell{0.064 \\ (0.00)} & \makecell{0.808  \\ (0.01)} & \makecell{0.099 \\ (0.01)} & \makecell{\textbf{0.871}  \\ \textbf{(0.00)}} & \makecell{\textbf{0.047} \\  \textbf{(0.01)}} \\
    
    \midrule
    \textbf{15 bins} \\
    \midrule

    Synthesis & \makecell{0.879 \\ (0.01)} & \makecell{0.049 \\ (0.01)} & \makecell{0.961 \\ (0.00)} & \makecell{0.026 \\ (0.00)} & \makecell{0.800 \\ (0.11)} & \makecell{0.224 \\ (0.08)} & \makecell{\textbf{0.904} \\ \textbf{(0.00)}} & \makecell{\textbf{0.04} \\ \textbf{(0.00)}} & \makecell{0.802 \\ (0.00)} & \makecell{0.096 \\ (0.01)} & \makecell{0.875 \\ (0.00)} & \makecell{0.052  \\  (0.00)} \\
    Synthesis+ & \makecell{0.881 \\ (0.01)} & \makecell{0.050 \\ (0.01)} & \makecell{\textbf{0.9605} \\ \textbf{(0.00)}} & \makecell{\textbf{0.024} \\ \textbf{(0.00)}} & \makecell{0.863 \\ (0.09)} & \makecell{0.203 \\ (0.10)} & \makecell{0.901 \\ (0.01)} & \makecell{0.055 \\ (0.01)} & \makecell{0.824 \\ (0.01)} & \makecell{0.087  \\  (0.01)	} & \makecell{0.879 \\ (0.01)} & \makecell{0.055  \\  (0.01)} \\
    
    \midrule
    \textbf{20 bins} \\
    \midrule

    Synthesis & \makecell{0.883 \\ (0.00)} & \makecell{0.046 \\ (0.01)} & \makecell{0.959 \\ (0.00)} & \makecell{0.027 \\ (0.00)} & \makecell{0.808 \\ (0.12)} & \makecell{0.180 \\ (0.07)} & \makecell{0.900 \\ (0.00)} & \makecell{0.048 \\ (0.00)} & \makecell{	0.818 \\  (0.01)} & \makecell{0.089 \\  (0.01)} & \makecell{0.871 \\  (0.01)} & \makecell{0.054  \\  (0.00)}  \\
    Synthesis+ & \makecell{\textbf{0.890} \\ \textbf{(0.00)}} & \makecell{\textbf{0.046} \\ \textbf{(0.01)}} & \makecell{0.959 \\ (0.00)} & \makecell{0.026 \\ (0.00)} & \makecell{\textbf{0.950} \\ \textbf{(0.04)}} & \makecell{\textbf{0.224} \\ \textbf{(0.03)}} & \makecell{0.896 \\ (0.01)} & \makecell{0.049 \\ (0.01)} & \makecell{\textbf{0.820}  \\ \textbf{(0.00)}	} & \makecell{\textbf{0.075}  \\  \textbf{(0.00)}	} & \makecell{0.867 \\  (0.01)} & \makecell{0.046 \\  (0.01)} \\
    % & \makecell{X \\ X} & \makecell{X \\ X} & \makecell{X \\ X} & \makecell{X \\ X} \\
    \bottomrule
  \end{tabularx}
  \caption{Model Performance and Calibration on Real Test Data. Highlighted values considered both ACC and ECE and weigh more on ECE.}
  \label{table:results}
\end{table*}

\subsection{Synthetic Data Generation}
Synthetic text generation is performed using version \textit{Llama-2-7b-chat-hf} of Llama 2 at a temperature $T=0.1$. We apply the two-stage and three-shot learning generation method proposed in the paper \cite{sahu-etal-2023-promptmix} to guarantee diversity and authenticity. First, we define each label and provide three examples for each one (Appendix \ref{appendix:c}). Then, we present the example text from the previous selection stage along with the predicted probability of this example that was extracted from the trained BERT$_{base}$ classifier. We then instruct llama 2 to act as the base classifier to generate three similar texts that could be classified with specific probability requirements. Next, we instruct llama 2 to relabel the generated texts to ensure them belong to the "correct" class. Table \ref{tab:prompt-example} illustrates the inputs, prompts, and outputs for generating synthetic data using the SE dataset. Additional prompts for different scenarios can be found in Appendix \ref{appendix:d}.

% \noindent\textbf{Sample Prompt.} Suppose an example text is from a overconfident and low probability area, and $D_m$ = 0.1, so our prompt to generate similar texts is: 
% \\
% % \textit{"""An example XXXX, which belongs $20\%$ to complaint and $80\%$ to not\_complaint (based on a classifier's categorization). Now I ask you act as that classifier and based on this example, generate a diverse set of 3 short utterances where each utterance belongs $10\%$ to complaint and $90\%$ to not\_complaint."""} 

% \fbox{\begin{minipage}{15em}\textit{"""An example XXXX, which belongs $20\%$ to complaint and $80\%$ to not\_complaint (based on a classifier's categorization). Now I ask you act as that classifier and based on this example, generate a diverse set of 3 short utterances where each utterance belongs $10\%$ to complaint and $90\%$ to not\_complaint."""} 
% \end{minipage}}
% \\

\subsection{Evaluation}
Results are assessed on real test set. \textbf{Baseline} refers to the results trained on the model in Step 1. Suppose we have a total of $N$ original data points in the training and validation set, and there are $S_{1}$ data points in target bins from the validation set. Let LLMs generate $S_{2}$ synthetic data points, and $S_{2}$ = $S_{1}$. \textbf{Synthesis} refers to the results that we retrain the model by replacing $S_{1}$ original data points with $S_{2}$ synthetic data points. \textbf{Synthesis+} indicates that we add $S_{2}$ synthetic samples into the original $N$ data points.

In addition to the baseline, we also compare the performance of our methods against some widely used model calibration techniques. \textbf{Isotonic regression} \cite{zadrozny2001obtaining} employs a non-parametric method that adjusts predicted probabilities to align with observed outcomes, and \textbf{Platt scaling} \cite{platt1999probabilistic} fits a logistic regression model to calibrate classifier scores based on predicted probabilities. \textbf{Monte Carlo dropout} \cite{gal2016dropout} randomly masks nodes to estimate the probability distribution. In our paper, the model makes 10 predictions for each instance and each time with a different dropout mask. \textbf{Temperature scaling} \cite{guo2017calibration} 
works by dividing the pre-softmax output by a temperature \textit{T} and the optimal value of \textit{T} is estimated by the validation dataset. All experiments use the same  BERT$_{base}$ model parameters.

% \textbf{Syn(fixVal)} means we keep the validation set the same as the baseline and only add synthetic data $S_{2}$ in the training set. 

\subsection{Results}

% To validate the calibration improvement, performance metric is computing based on real test data (Table \ref{table-dataset}). See model performance based on complaints data(TC) in table \ref{table-results}. \textcolor{blue}{will update other dataset result later.} \textbf{Syn(val)} means the target bins in calibration reliability diagram are based on validation set and when retrain the model, add synthetic data into original training and validation set, re-split them into new train and validation set; \textbf{Syn(valFixed)} means the target bins in calibration reliability diagram are based on validation set and when retrain the model, fix the original validation set, add synthetic data only into the training set. \textbf{Syn(train)} denotes the target bins in calibration reliability diagram are based on the training and validation set; \textbf{Syn(trainMixup)} denotes the target bins in calibration reliability diagram are based on the training and validation set and add some points near decision boundary (i.e., 45\% of class 0 and 55\% of class 1) for wrong predictions.  

We run each experiment for three random seeds and report the average value (with standard deviation in brackets) of accuracy and ECE in Table \ref{table:results}. 
 By adding synthetic data with a size of 7\%-18\% \footnote{The validation set is used to identify poorly calibrated bins. We set a predefined threshold of 0.03, and only bins with gaps exceeding this threshold are selected.}
 of the training set, we would have a 21-33\% ECE decrease. Taking both accuracy and ECE into account, our synthetic data replacement (synthesis) and synthetic data add-on (synthesis+) methods outperform other calibration approaches in five out of six datasets. Temperature scaling can sometimes achieve lower ECE, but a key disadvantage is that it doesn't affect accuracy. On the other hand, while dropout can improve model calibration, it carries the risk of reducing accuracy, as seen in the results on the Arxiv-10 dataset. We also observe a clear association of a larger bin number with lower ECE\footnote{We use the bin's average confidence to represent each instance within that bin, so having more bins could lead to more accurate probability estimates in synthetic data generation.}. In addition, the results of our approach have smaller variances compared with those of the baseline. 

Whether imbalanced datasets (TC and SE) or balanced datasets (SUBJ, B77, Arxiv, and Medical), improvements in uncertainty calibration are fairly comparable on average. It is also shown that even though the baseline model for the SUBJ dataset already has outstanding accuracy, our approach can still make the model better calibrated without degrading the model's classification performance. Results from B77 have a larger variance due to its smaller data size. 

% For two imbalanced datasets (TC and SE), both synthetic data replacement (synthesis) and synthetic data added on (synthesis+) can improve model accuracy and lower the expected calibration error. 
% Even though increasing the sample size is more likely to attain a lower ECE, the best improvement of ECE is not the one with the most synthetic data added on. Quality data generation and find the important instances could be the key. 

% It is also shown that even though the baseline model for SUBJ dataset already has an outstanding accuracy, our approach can still make the model better calibrated without degrading the model's classification performance. Results from B77 have a larger variance due to its smaller data size. 

\subsection{Ablation Study}
To discuss if the LLM's self-calibration capability strongly impacts our approach, we instruct \textit{Llama-2-7b-chat-hf} with few-shot learning and set $top_{k}$ = 1 in which we obtain the conditional probability of one class $P(label|text)$. Then we computer the accuracy and ECE from LLMs.

\begin{table}[ht]
  \centering
  \scriptsize % Reduced font size
  \setlength{\tabcolsep}{4pt} % Adjust horizontal space between columns
  \begin{tabular}{@{}lcccc@{}}
    \toprule
    & \textbf{$\text{LLM}_{ACC}$} & \textbf{$\text{LLM}_{ECE}$} & \textbf{$\text{Syn}_{ACC}(\%)$} & \textbf{$\text{Syn}_{ECE}(\%)$}\\
    \midrule
    \textbf{$\text{LLM}_{ACC}$}   & 1 & -0.737   & 0.592 & -0.566 \\
    \textbf{$\text{LLM}_{ECE}$} &  -0.737 & 1  & -0.026 & 0.423 \\
    \bottomrule
  \end{tabular}
  \caption{Pearson Correlation Coefficient on six datasets. $\text{Syn}_{ACC}(\%)$ and $\text{Syn}_{ECE}(\%)$  denotes the percentage of the downstream model's accuracy increases and how much percentage of the expected calibration error is decreased respectively. }
  \label{table-ablation}
\end{table}

In Table \ref{table-ablation}, we didn't observe a strong negative correlation coefficient between $\text{LLM}_{ACC}$ and $\text{Syn}_{ECE}(\%)$, indicating there is no empirical evidence that shows the calibration ability of LLMs determines the application of our proposed methods. Additionally, we found a moderate positive association between the llama’s accuracy and the accuracy improvement in downstream tasks. This suggests that the prediction accuracy of LLMs, rather than calibration capability, plays a more important role in downstream models' performance. Therefore, using advanced LLMs (such as Llama 3.2) or fine-tuning LLMs to incorporate domain knowledge could yield better performance when applying our approach.

% Here are the results. syn_acc(\%) denotes the percentage of the downstream model's accuracy increases and syn_ece(\%) means how much percentage of the expected calibration error is decreased by employing our proposed method.

% ablation 
% syn(valFixed)   & & 0.893 & 0.0434 & & & & & & & & & \\
% syn(trainMixup) & & 0.948 & 0.0378 & & & & & & & & & \\
    
% SentiEval: 
% original: ACC: 0.87 ece: 0.0655
% syn-val-bin10-cf: ACC: 0.907, ece: 0.0408 - n_syn: 18
% syn-val-bin10-cf_more: ACC: 0.875, ece: 0.0651, 0.0408 - n_syn: 18 +14
% 

% SubJ: 
% original: ACC: 0.961, ECE: 0.0285
% 
% 

%sample size (1: 1, 1:2) , target train or only validation  

% Ablation experiments
% only add synthetic wrong instances  - compare 
% 

\section{Related Work}

Model calibration has emerged as an open challenge in machine learning as concerns arise regarding the responsible and ethical use of ML-enabled systems. Several methods have been proposed, including Platt Scaling \cite{platt1999probabilistic}, Isotonic regression \cite{zadrozny2001obtaining}, among others. Both of them somewhat change the predicted probability, which could lower the predicted accuracy. In the computer vision field, a mixup method \cite{zhang2017mixup} has been proposed to overcome the shortcomings of data scarcity. It combines two instances from the original dataset with different proportions. A follow-up paper \cite{wen2020combining} investigates the computer vision task calibration by using the mixup approach and concludes that this approach could impair the model calibration. However, the calibration issue in the NLP field has rarely been discussed. 

Several benefits from using synthetic data have been explored in \cite{sahu-etal-2023-promptmix}. It is found that ML prediction accuracy can be improved significantly by adding synthetic data generated near the decision boundary. On the other hand, a recent paper \cite{li2023synthetic} investigates on potential and limitations of synthetic data generated from LLM for text classification tasks and concludes that while synthetic data can be beneficial in certain scenarios, it does not consistently enhance model performance. Our research is different from theirs in that we provide a strategy that enables both good generalization and uncertainty calibration.

% Because the classifiers improve their ability on more hard cases, they would perform pretty well on easier cases. Also, the data diversity brought by synthetic data results in superior generalizability and makes the ML model more robust. Our approach is different from theirs in that we provide a strategy that enables good generalization and uncertainty calibration at the same time. 

\section{Conclusion} % or discussion

In the era of large models, we believe smaller models still hold tremendous values in, e.g., edge computing and specialized downstream machine learning tasks. We derive the expected calibration error bound for ML models and explore the possibility of leveraging synthetic data to mitigate calibration error. Through empirical validation with text classification tasks, we demonstrate the usefulness of our method; that is, by harnessing the power of LLMs, purposefully generated synthetic data can be utilized to train smaller downstream NLP tasks, achieving both strong classification performance and calibration error reduction.

\section*{Limitations and Future Work}

%  medium-sized LLM like 75b model or 
% multi-class calibration 

While increasing the sample size generally helps reduce the Expected Calibration Error (ECE), simply adding more synthetic data may not always lead to optimal model performance, as excessive synthetic data can cause overfitting. Therefore, the focus should be on generating high-quality data and strategically identifying instances that require better calibration. It's also important to note that the primary objective of this paper is not to compare or evaluate the capabilities of Large Language Models (LLMs). Rather, we assume that an updated, optimally performing LLM could generate higher-quality synthetic data, which could, in turn, enhance the accuracy of downstream tasks and improve model calibration using our proposed methodology.

%Even though increasing the sample size is more likely to attain a lower ECE, the optimal improvement in ECE does not necessarily stem from adding the most synthetic data. Excessive synthetic data could lead to overfitting. Quality data generation and a strategic approach to identifying instances requiring calibration are likely key factors. However, it is important to note that the primary objective of this paper is not to compare or evaluate the capabilities of LLMs. Instead, we operate under the assumption that data generated from an updated version of LLMs exhibiting optimal performance would facilitate better synthetic text data generation. Subsequently, this has the potential to enhance the accuracy of downstream tasks and model calibration based on our methodology.

In our experiments, we applied a 0.03 threshold to filter out ill-calibrated bins, leaving room for future work to investigate how varying cutoff values might influence calibration enhancement. While our method focuses on text classification applications, there is potential to extend this approach to other downstream NLP tasks. Additionally, future research could explore the use of generative models beyond Large Language Models (LLMs), broadening the scope of applicability. Finally, extending our method to multi-class classification models is proposed as an area for future work.

% \section*{References}
% \bibliographystyle{acl_natbib}
\bibliography{refs}

\clearpage
\appendix
\section{Appendix} % or discussion

\label{appendix:a}
Our code is implemented based on Pytorch 2.2.1 and the pre-trained Bert\textsubscript{base} model is downloaded from the huggingface library. Both llama2 and Bert\textsubscript{base} run on Nvidia P100 GPUs. 

% Adam optimizer

\begin{table}[h]
  \centering
  \begin{tabular}{ll}
    \toprule
      & Parameters  \\
    \midrule
    optimizer & Adam \\
    max length & 512 \\
    embedding dim & 768 \\
    batch size & 32\\
    learning rate & 1e-6    \\
    dropout&  0.5 \\
    epoch & 5\\
    \bottomrule
  \end{tabular}
   \caption{Model Parameters (Bert\textsubscript{base})}
  \label{table-model}
\end{table}

% \rule{8cm}{0.4pt} % 5cm wide, 0.4pt thick

\lstset{
    basicstyle=\ttfamily\footnotesize,  % Reduce font size
    breaklines=true,                     % Enable line breaking
    columns=fullflexible,                % Make it use the full column width
    keepspaces=true                      % Keep indentation spaces
}

We provide the code used to generate the corresponding prompts based on the scenarios to which the bins in Table \ref{table-synGen} belong.

\rule{7cm}{0.4pt} % 5cm wide, 0.4pt thick

\begin{lstlisting}
def gen_prompt(conf, diff, label0, label1, indicator):
    """ diff: the gap against the perfect calibration line.
        indicator: based on table 1, which the bin belongs to.
        revised_conf: set a bound if the gap is too large 
        that makes the generated instances assigned correct labels. """
    
    if 'low' in indicator and 'under' in indicator: 
        revised_conf = 45 if conf + diff >= 50 else conf + diff
        generation_prompt = f"which belongs {100-conf}% to {label0} and {conf}% to {label1} 
        (based on a classifier's categorization). Now I ask you to act as that classifier 
        and based on this example, generate a diverse set of 3 short utterances where each 
        utterance belongs {100-revised_conf}% to {label0} and {revised_conf}% to {label1}: "

    if 'low' in indicator and 'over' in indicator: 
        revised_conf = 5 if conf - diff <= 0 else conf - diff
        generation_prompt = f"which belongs {100-conf}% to {label0} and {conf}% to {label1} 
        (based on a classifier's categorization). Now I ask you to act as that classifier 
        and based on this example, generate a diverse set of 3 short utterances where each 
        utterance belongs {100-revised_conf}% to {label0} and {revised_conf}% to {label1} 
        (no explanation):"

    if 'high' in indicator and 'under' in indicator: 
        revised_conf = 95 if conf + diff >= 100 else conf + diff
        generation_prompt = f"which belongs {100-conf}% to {label0} and {conf}% to {label1} 
        (based on a classifier's categorization). Now I ask you to act as that classifier 
        and based on this example, generate a diverse set of 3 short utterances where each 
        utterance belongs {100-revised_conf}% to {label0} and {revised_conf}% to {label1} 
        (no explanation):"

    if 'high' in indicator and 'over' in indicator: 
        revised_conf = 55 if conf - diff <= 50 else conf - diff
        generation_prompt = f"which belongs {100-conf}% to {label0} and {conf}% to {label1} 
        (based on a classifier's categorization). Now I ask you to act as that classifier 
        and based on this example, generate a diverse set of 3 short utterances where each 
        utterance belongs {100-revised_conf}% to {label0} and {revised_conf}% to {label1} 
        (no explanation):"
        
    return generation_prompt

\end{lstlisting}

\clearpage

% \clearpage

% \onecolum
% \section{Appendix}
% \label{appendix:b}

\onecolumn
\section{Appendix}
\label{appendix:b}
\textbf{The Expected Calibration Bound Proof}: \\
From equation (\ref{eq:ac-conf}) in section 2, we extend the definition of accuracy and confidence from bin-wise to data-wise: 

% \begin{table*}[h]
% \centering
\small
\[
\text{Acc}(X) = \sum_{m=1}^M \frac{|B_m|}{n} \text{Acc}(B_m), \quad \text{Conf}(X) = \sum_{m=1}^M \frac{|B_m|}{n} \text{Conf}(B_m)
\]
\normalsize
correspondingly,
\small
\[
\text{Acc}(X^*) = \sum_{m=1}^M \frac{|B_m^*|}{n} \text{Acc}(B_m^*), \quad \text{Conf}(X^*) = \sum_{m=1}^M \frac{|B_m^*|}{n} \text{Conf}(B_m^*)
\]
% \end{table*}
\normalsize
According to Hoeffding’s inequality, we have: 
\small
$$
P(\left| Acc(X) - Acc(X^*) \right| >\epsilon_a ) \leq 2 \exp(-2 \epsilon_a^2 n). 
$$
\normalsize
where, $Acc(X)$ means the expected accuracy in the model; $Acc(X^*)$ is the observed accuracy of training data. $\epsilon_a$ is the error for accuracy and we let $\delta_a = 2 \exp(-2 \epsilon_a^2 n)$. The derivation from the left side of the inequality:  
% \begin{table*}
\small
\begin{flalign*}
\begin{split}
 &P\left(\left| \text{Acc}(X) - \text{Acc}(X^*) \right| > \epsilon_a \right) \\
= &P\left(\left| \sum_{m=1}^M \frac{|B_m|}{n} \text{Acc}(B_m) - \sum_{m=1}^M \frac{|B_m|}{n} \text{Acc}(B_m^*) \right| > \epsilon_a \right) \\
= &P\left(\sum_{m=1}^M \frac{|B_m|}{n}\left| \text{Acc}(B_m) - \text{Conf}(B_m) + \text{Conf}(B_m) + \text{Conf}(B_m^*) - \text{Conf}(B_m^*) - \text{Acc}(B_m^*) \right| > \epsilon_a \right) \\
= &P\left(\sum_{m=1}^M \frac{|B_m|}{n}\left| \text{Acc}(B_m) - \text{Conf}(B_m) - [\text{Acc}(B_m^*) - \text{Conf}(B_m^*)] + \text{Conf}(B_m)  - \text{Conf}(B_m^*) \right| > \epsilon_a \right)\\
\geq &P\left(\sum_{m=1}^M \frac{|B_m|}{n}\left| \text{Acc}(B_m) - \text{Conf}(B_m) - [\text{Acc}(B_m^*) - \text{Conf}(B_m^*)] \right| - \sum_{m=1}^M \frac{|B_m|}{n}\left| \text{Conf}(B_m) - \text{Conf}(B_m^*) \right| > \epsilon_a \right) \\
= &P\left(\sum_{m=1}^M \frac{|B_m|}{n}\left| \text{Acc}(B_m) - \text{Conf}(B_m) - [\text{Acc}(B_m^*) - \text{Conf}(B_m^*)] \right| > \epsilon_a +  \sum_{m=1}^M \frac{|B_m|}{n}\left| \text{Conf}(B_m) - \text{Conf}(B_m^*) \right| \right) \\
\geq &P\left(\sum_{m=1}^M \frac{|B_m|}{n}\left| \text{Acc}(B_m) - \text{Conf}(B_m) \right| - \sum_{m=1}^M \frac{|B_m|}{n}\left| \text{Acc}(B_m^*) - \text{Conf}(B_m^*) \right| > \epsilon_a +  \left|\text{Conf}(X) - \text{Conf}(X^*) \right| \right) \\
= &P\left( \text{ECE}(X)  - \text{ECE}(X^*)  > \epsilon_a +  \left|\text{Conf}(X) - \text{Conf}(X^*) \right| \right) 
\end{split}
\end{flalign*}

% \end{table*}
\normalsize
Combined with the right side of the inequality:  
% \begin{table*}
% \small
\begin{flalign*}
&P\left( \text{ECE}(X)  - \text{ECE}(X^*)  > \epsilon_a +  \left|\text{Conf}(X) - \text{Conf}(X^*) \right|  \right) \leq 2 \exp(-2 \epsilon_a^2 n) \\
= &P\left(|\text{ECE}(X)  - \text{ECE}(X^*)|  > \epsilon_a +  \left|\text{Conf}(X) - \text{Conf}(X^*) \right|  \right) \leq 4 \exp(-2 \epsilon_a^2 n) \\
= &P\left( \left|\text{ECE}(X)  - \text{ECE}(X^*) \right| > \epsilon_{ECE} \right) \leq 4 \exp(-2 \epsilon_a^2 n) \\
= &P\left( \left|\text{ECE}(X)  - \text{ECE}(X^*) \right| > \epsilon_{ECE} \right) \leq 4 \exp(-2 (\epsilon_{ECE} - \left|\text{Conf}(X) - \text{Conf}(X^*) \right| )^2 n)
\end{flalign*}
% \end{table*}

where $ \epsilon_{ECE} =  \epsilon_a + \left|\text{Conf}(X) - \text{Conf}(X^*) \right| = \epsilon_a + \sum_{m=1}^M \frac{|B_m|}{n}\left| \text{Conf}(B_m) - \text{Conf}(B_m^*)\right|$.
% ] 
\\

\clearpage

% \section{Appendix}
% \label{appendix:c}

\onecolumn
\section{Appendix}
\label{appendix:c}

\begin{table*}[h!]
  \centering
  \small
  \begin{adjustbox}{max width=\textwidth} % Adjusts the table to the maximum width of the text
  \begin{tabular}{l m{15cm}} % Adjust the width of the second column as needed
    \toprule
    \multicolumn{2}{c}{System Prompt}  \\ % Centered heading across two columns
    \midrule
    TC &  \makecell[l]{Consider the task of classifying between the following classes (along with some examples):\\
        1. complaint, which is about customer inquiries on a state of affairs, product, organization or event 
        to express a negative \\mismatch between reality and expectations.
        Some examples of utterances include: \\
            - Dear @nvidia, I don't think I should have to roll back to driver v270.61 to make my games work, and my desktop \\not glitch out.\\
            - {@}FC\_Help hi m order is 913181 did you revise the money? if you did.. how about the shipping ?\\
            - {@}FC\_Help Will you be getting the wendy cotton v neck dress in pavlova back in stock on the site?\\
            \\
        2. not\_complaint, which is the opposite of complaint mentioned above, about customer regular or normal inquiries on a \\state of affairs, product, organization or event  without any expression related to a negative mismatch between reality and \\expectations.
        Some examples of utterances include:\\
            - @FC\_Help How can I get a hold of you so we can discuss the problem I am having with my coat?\\
            - @FC\_Help I need to check my order.\\
            - @FC\_Help looking for "bright carol" or "stained glass" dress. do you have these in stock anymore?} \\
    \midrule
    SUBJ & \makecell[l]{Consider the task of classifying between the following classes (along with some examples):\\
        1. objective, which is assigned to text that presents factual information, descriptions, or statements without personal \\opinions, emotions, or bias. It focuses on delivering \\facts or information that is independent of the writer's personal feelings or beliefs.
        Some examples of utterances include:\\
            - "nicklas passes out , and the next day when he returns to school he notices that nobody seems to notice him."\\
            - "when reuben buys a black-market cure for his unusual chest complaint, jenny is forced to make a terrible sacrifice."\\
            - "raj has always had a unrequited childhood crush on a friend named tina, but tina's best friend pooja has always had a \\crush on raj."\\
            \\
        2. subjective, which is applied to text that expresses personal opinions, feelings, beliefs, or thoughts. It often includes \\evaluative language, personal experiences, or interpretations, reflecting the writer's personal stance or emotional reaction.\\
        Some examples of utterances include:\\
            - "for its seriousness, high literary aspirations and stunning acting, the film can only be applauded."\\
            - "an inelegant combination of two unrelated shorts that falls far short of the director's previous work in terms of both \\thematic content and narrative strength."\\
            - "what's needed so badly but what is virtually absent here is either a saving dark humor or the feel of poetic tragedy."} \\
    \midrule
    B77 & \makecell[l]{Consider the task of classifying between the following classes (along with some examples):\\
        1. age\_limit, which is about customer inquiries on age-related restrictions for opening a bank account.\\
        Some examples of utterances include:\\
            - Can I get an account for my son?\\
            - Can my teenager have an account?\\
            - How young can I be to open my own account?\\
            \\
        2. atm\_support, which is about users asking how to use an ATM, where to find one, or any other clarifications about\\ a transaction at an ATM. Some examples of utterances include:\\
            - Is the closest ATM to me within 2 miles?\\
            - Are there only certain ATM machines where I can use this card?\\
            - Do you know the closest ATM?}  \\
    \midrule
    SE  & \makecell[l]{Consider the task of classifying between the following classes (along with some examples):\\
        1. negative, which is assigned to content that expresses negative feelings, emotions, or attitudes. Examples include \\statements of dissatisfaction, sadness, anger, or criticism.
        Some examples of utterances include:\\
            - "even with newborn diapers it filled way too fast."\\
            - "bluetooth does not work on this phone."\\
            - "also, some other mp3 players such as the nitrus allow you to play wma (windows media audio) files, whereas the \\ipod does not."\\
            \\
        2. positive, which is applied to content that expresses positive feelings, emotions, or attitudes. Examples include \\statements of happiness, satisfaction, praise, or optimism.\\
        Some examples of utterances include:\\
            - "4 megapixels is enough for anybody and the photo quality is awesome."\\
            - "an extra gig of room, fm radio, voice recorder, pim applications with sync to outlook."\\
            - "all the buttons \& necessary lil' gadgets are on the sides of the player which is nice for when you are holding it in \\the palm of your hand."} \\ 
    \bottomrule
  \end{tabular}
  \end{adjustbox}
  \caption{System Prompt for Data Generation}
  \label{table:sysprompt}
\end{table*}

\clearpage

% \twocolumn
% \section{Appendix}

\onecolumn
\section{Appendix}
\label{appendix:d}

\begin{table*}[!h]
    \centering
    \small
    \adjustbox{max width=\textwidth}{ % Adjust the table to fit within the text width
    \begin{tabular}{|c |p{1.6cm} |p{3cm} |p{4cm}| p{3cm}| p{1.2cm} |}
        \hline
        \textbf{Dataset} & \multicolumn{2}{c|}{\textbf{Input}} & \centering \textbf{Prompt} & \textbf{\thead{Generated Text\\ ($x^{syn}_{i}$ , $\hat{P}(y_{i}|x^{syn}_{i})$)}} & \textbf{\thead{Label \\(Y)}} \\ \hline

        \multirow{11}{*}{\textbf{TC}} & \centering  \multirow{9}{*}{$x_{i}$} & \vspace{6mm} @UbisoftSupport When will u guys fix the jager glitch? \newline \newline \newline & \vspace{2mm} \multirow{2}{4cm}{An example $x_i$ which belongs 75\% to complaint and 25\% to not\_complaint (based on a classifier's categorization). Now I ask you act as that classifier and based on this example, generate a diverse set of 3 short utterances where each utterance belongs 65\% to complaint and 35\% to not\_complaint.}  & \vspace{4mm} \multirow{5}{3cm}{Hey @UbisoftSupport, I'm still experiencing the jager glitch and it's really affecting my gameplay. Can you please fix it asap? (65\% complaint, 35\% not\_complaint)} &  \multirow{11}{*}{complaint}\\ \cline{2-3}
        
        & \centering \multirow{2}{*}{$\hat{P}(y_{i}|x_{i})$ }& \centering \multirow{2}{*}{ 0.75} \newline & &  &  \\ \cline{2-3}
                     
        & \centering miscalibration type &  \centering \multirow{2}{*}{ overconfidence}  \newline & & & \\ \hline

        \multirow{11}{*}{\textbf{SUBJ}} & \centering  \multirow{9}{*}{$x_{i}$} & \vspace{7mm} the gripping story of the dutch bettien martens , based on true facts. \newline \newline \newline& \vspace{2mm} \multirow{2}{4cm}{An example $x_i$ which belongs 95\% to objective and 5\% to subjective (based on a classifier's categorization). Now I ask you act as that classifier and based on this example, generate a diverse set of 3 short utterances where each utterance belongs 70\% to objective and 30\% to subjective.}  & \vspace{4mm} \multirow{5}{3cm}{The remarkable tale of Bettien Martens serves as a testament to the enduring nature of hope and determination, even in the face of unimaginable hardship. (70\% objective, 30\% subjective)} &  \multirow{11}{*}{objective}\\ \cline{2-3}
        
        & \centering \multirow{2}{*}{$\hat{P}(y_{i}|x_{i})$ }& \centering \multirow{2}{*}{ 0.10} \newline & &  &  \\ \cline{2-3}
                     
        & \centering miscalibration type &  \centering \multirow{2}{*}{ underconfidence}  \newline & & & \\ \hline
        
        \multirow{11}{*}{\textbf{B77}} & \centering  \multirow{9}{*}{$x_{i}$} & \vspace{10mm} Can my teenager have an account? \newline \newline  & \vspace{2mm} \multirow{2}{4cm}{An example $x_i$ which belongs 65\% to age\_limit and 35\% to atm\_support (based on a classifier's categorization). Now I ask you act as that classifier and based on this example, generate a diverse set of 3 short utterances where each utterance belongs 85\% age\_limit and 15
        \% to atm\_support.}  & \vspace{7mm} \multirow{5}{3cm}{Can my toddler open their own bank account? (85\% age\_limit and 15\% to atm\_support)} &  \multirow{11}{*}{age\_limit}\\ \cline{2-3}
        
        & \centering \multirow{2}{*}{$\hat{P}(y_{i}|x_{i})$ }& \centering \multirow{2}{*}{ 0.65} \newline & &  &  \\ \cline{2-3}
                     
        & \centering miscalibration type &  \centering \multirow{2}{*}{ underconfidence}  \newline & & & \\ \hline

        \multirow{11}{*}{\textbf{SE}} & \centering  \multirow{9}{*}{$x_{i}$} & \vspace{0.5mm} The zoom function on this camera is so loud that sometimes you will be unable to use it if you find yourself in a situation where you must be quiet.  & \vspace{2mm} \multirow{2}{4cm}{An example $x_i$ which belongs 75\% to negative and 25\% to positive (based on a classifier's categorization). Now I ask you to act as that classifier and based on this example, generate a diverse set of 3 short utterances where each utterance belongs 55\% to negative and 45\% to positive.}  & \vspace{4mm} \multirow{5}{3cm}{I love how this router can handle a large network, but the price is a bit steep for my taste. (55\% negative, 45\% positive)} &  \multirow{11}{*}{negative}\\ \cline{2-3}
        
        & \centering \multirow{2}{*}{$\hat{P}(y_{i}|x_{i})$ }& \centering \multirow{2}{*}{ 0.25} \newline & &  &  \\ \cline{2-3}
                     
        & \centering miscalibration type &  \centering \multirow{2}{*}{ overconfidence}  \newline & & & \\ \hline
    \end{tabular}
    }
    \caption{An example of generating synthetic data via LLM. \textbf{Input} contains the original text ($x_{i}$) and the average predictive probability of the bin it comes from ($\hat{P}(y_{i}|x_{i})$). \textbf{Generated Text} is the one after the relabeling process. \textbf{Note}: during re-fine tuning of the downstream model, we exclude $\hat{P}(y_{i}|x^{syn}_{i})$ and retain only $x^{syn}_{i}$ in the dataset.}
    \label{tab:prompt}
\end{table*}

% \label{appendix:d}

\twocolumn
% \section{Appendix}
% \subfile{sections/Appendix-5}

%\section{NeurIPS Paper Checklist}
% \subfile{sections/checklist}

\end{document}